\documentclass[11pt]{article}
\usepackage[preprint]{acl}
\usepackage{times}
\usepackage{latexsym}
\usepackage[T1]{fontenc}
\usepackage[utf8]{inputenc}
\usepackage{microtype}
\usepackage{booktabs}
\usepackage{amsmath,amssymb}
\usepackage{graphicx}
\usepackage{multirow}
\newcommand{\ptrue}{P(\mathrm{True})}

\title{Disentangling Question, Response, and Verifier Languages\\
in Multilingual P(True) Self-Evaluation}
\author{
 \textbf{Jewon Yeom}\thanks{\texttt{jewon0908@snu.ac.kr}},
 \textbf{Jaewon Sok},
 \textbf{Seonghyeon Park},
 \\
 \textbf{Jeongjae Park},
 \textbf{Hwiyeong Lee},
 \textbf{Taesup Kim}\thanks{Corresponding author: \texttt{taesup.kim@snu.ac.kr}}
\\[4pt]
 Graduate School of Data Science, Seoul National University
}

\begin{document}
\maketitle

\begin{abstract}
When a language model is asked whether its own answer is correct,
three languages are involved: that of the question, that requested
for the answer, and that of the verifier instruction asking for the judgment.
Multilingual confidence evaluations seldom separate the three.
We vary them independently in a six-language factorial design with ten
open-weight models (3B--14B) on LSQA, MMMLU and TriviaQA, holding each
answer fixed across verifier languages and measuring confidence with a
first-token P(True) probe. The three languages move different parts of
the evaluation: which of them governs accuracy depends on the benchmark,
matching the response language to the question or verifier language
sharpens discrimination but not calibration, and the verifier language
alone shifts calibration error by up to 0.04 without changing accuracy.
The scorer matters as well: an unvalidated judge would have
shrunk the MMMLU English--Hindi accuracy gap from 0.198 to 0.038.
Multilingual confidence results, model rankings included, are therefore
comparable only when all three languages are reported and the scorer
is validated.
\end{abstract}

\section{Introduction}
A multilingual evaluation of model confidence involves at least three
languages: that of the question, that requested for the answer, and
that of the instruction that elicits confidence. Applications mix
them, as when an English probe audits answers to Hindi questions, yet
evaluations seldom report them separately
\citep[e.g.,][]{krause2023confidently,xue2024mlingconf}, although each
is known to matter \citep{shi2023language,marchisio2024understanding,behzad2024ask}.
If each language moves
accuracy, discrimination or calibration in its own way, results from
different language settings are not comparable.

\begin{figure}[t]
\centering
\includegraphics[width=\columnwidth]{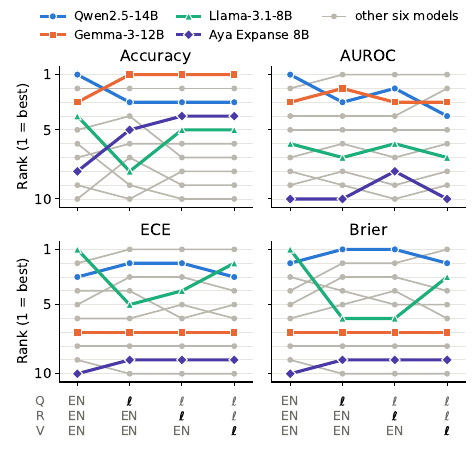}
\caption{Model ranks as the question ($Q$), requested response ($R$)
and verifier ($V$) languages switch one by one from English to the
same $\ell$ (mean over five $\ell$ and three benchmarks). Six model pairs
swap accuracy order from first to last setting (95\% bootstrap;
Appendix~\ref{app:additional}).}
\label{fig:teaser}
\end{figure}

We therefore vary three languages independently: (a) the question
language $Q$, (b) the requested response language $R$, and (c) the
verifier language $V$, that of the instruction asking whether
an answer is correct. Each generated response is held
fixed across verifier languages, so $V$ can change confidence but not
accuracy. We use the first-token $\ptrue$ probe
\citep{kadavath2022language}, the probability a model assigns to
``True'' for its own answer, because all three languages enter its
prompt; we call it \emph{verifier confidence}. We ask which language
contrasts affect accuracy, AUROC, ECE and Brier score per
benchmark, and whether discrimination and calibration move
together.

The three languages behave as experimental variables: switching
them from English one at a time reorders the ten models we evaluate
(Figure~\ref{fig:teaser}), and their effects differ by benchmark and
by metric (Figure~\ref{fig:effects}). The three
benchmarks differ in content, provenance and response format, so
their comparison is informative but does not identify a causal
effect of task regime or response format.

\textit{In short, each of the three languages moves a different part of
the evaluation}, so no single language setting can stand in for the
others. Our contributions are as follows.
(1) A six-language factorial design that holds each generated
response fixed across verifier languages, applied to ten open-weight
models on three benchmarks.
(2) \textit{The benchmark decides which language governs accuracy.}
On MMMLU the question language separates accuracy far more than the
requested response language; on LSQA the item's content subset and
the requested response language dominate; on TriviaQA both languages
matter.
(3) \textit{Discrimination and calibration respond differently.}
Matching the response language to the question or verifier language
raises AUROC on LSQA and TriviaQA in nearly every model but improves
calibration on neither, and the verifier language shifts ECE without
changing accuracy.
(4) \textit{The scorer can change the conclusions.} Human review
showed that the automated judge over-credits multiple-choice answers:
under its labels the MMMLU English--Hindi question gap is 0.038
instead of 0.198 and requests for English responses rank last instead
of first, so we score MMMLU by exact option match. Answers given
outside the requested language also account for part of TriviaQA's
English-question advantage. We therefore recommend reporting all three
languages, the realized response language and a per-benchmark
validation of the scorer.

\section{Related Work}
\paragraph{Confidence estimation.} \citet{kadavath2022language}
introduced $\ptrue$: a model proposes an answer, is asked whether
that answer is true, and the probability it assigns to ``True''
serves as its confidence. Other methods elicit verbalized confidence
\citep{lin2022teaching,tian2023just,xiong2024can}, measure semantic
consistency across samples \citep{farquhar2024detecting}, or rely on
extended reasoning \citep{yoon2025reasoning}; \citet{geng2024survey}
survey the area. Because our probe reads a single next-token logit,
it inherits the known gap between first-token probabilities and
generated answers \citep{wang2024answer,wang2024look}, and our claims
are restricted to this protocol. Multiple-choice evaluation is also
sensitive to option order and labels
\citep{zheng2024large,pezeshkpour2024large}, so we score MMMLU by the
generated option and analyze it apart from the free-text benchmarks.

\paragraph{Multilingual confidence and knowledge.}
\citet{xue2024mlingconf} study confidence estimation across languages
and distinguish language-agnostic from language-specific tasks.
Multilingual encoders are worst calibrated for languages far from
English \citep{ahuja2022calibration}; \citet{yang2024can} study the
calibration of multilingual question answering models,
\citet{zhou2025layerwise} use intermediate layers to improve it, and
\citet{krause2023confidently} find that answers to Hindi and Amharic
TriviaQA questions are not hedged even as accuracy drops. These
studies vary the prompt language as a whole, so all three
languages change together. Answers to the
same fact also differ across languages \citep{qi2023cross,wang2025lost},
intermediate layers of Llama-2 favor English
even for non-English input \citep{wendler2024llamas}, models reason as
well or better in English \citep{shi2023language,etxaniz2024multilingual},
and making a reasoning model think in the user's language can cost
accuracy \citep{qi2025crosslingual}. Models also often answer in a
language other than the requested one \citep{marchisio2024understanding},
and translated test sets carry translation artifacts
\citep{artetxe2020translation} and cultural bias
\citep{singh2025global}. We therefore vary the three languages
separately, label the realized response language and audit the
translated items.

\paragraph{Automated judges.} Model-based judges score free-text
outputs \citep{zheng2023judging,liu2023geval,kim2024prometheus2},
and M-Prometheus provides open multilingual judges
\citep{pombal2025mprometheus}. Their agreement with humans varies
across tasks \citep{bavaresco2024llms} and languages
\citep{hada2024metal}, and LLM judges give inconsistent verdicts on
parallel content across languages
\citep{fu2025reliable,fu2026languages,yin2026judge,mahmood2026rank}. We therefore validate our judge against
language-qualified reviewers and replace it where it fails
(Section~\ref{sec:human}).

\section{Experimental Design}
\subsection{Benchmarks and Languages}
\label{sec:benchmarks}
We use English (EN), French (FR), Chinese (ZH), Japanese (JA), Korean
(KO) and Hindi (HI); tables and figures use these ISO 639-1 codes.
Each benchmark contains 960 base items, with parallel question
versions in the six languages.

\paragraph{LSQA.}
LSQA follows the language-specific task design of
\citet{xue2024mlingconf}, but its items are new: they do not appear in
the public release of that work beyond a few generic questions. The
release contains 160 items per content-language subset, 40 in each of
the history, geography, social/culture and celebrity categories. For
each subset, questions and reference answers were generated in the
subset language and in English with Claude Opus~4.7
\citep{anthropic2026opus47} and checked by the
authors; the other four language versions were translated with the
same model. Responses
are concise free text. We retain the distinction between the cultural
subset and the language in which an item is asked.

\paragraph{MMMLU.}
We construct a deterministic selection from the public multilingual
release \citep{openai2024mmmlu} of MMLU \citep{hendrycks2021mmlu},
whose localized versions were produced by professional human
translators, with seed 3407 and 240 items in each of four subject groups:
STEM, humanities, social sciences, and other. The expected response is an option letter with a brief explanation
in the requested language. An option letter alone does not reveal
the realized response language. The gold option is the answer
letter, which the six versions share, and correctness is scored
by exact option match (Section~\ref{sec:labels}). The human
translation audit (Section~\ref{sec:human}) found that localized
MMMLU items carry most of the observed defects, mainly mistranslated
or corrupted options, which exact match cannot separate from model
errors. The release has no shared item identifier, so we match the
six versions of an item by its row position within each subject.
Before sampling we excluded 238 of the 14,042 source rows: one
subject whose localized rows do not line up with the English ones
(high-school world history) and every row whose answer letter
differs from the English version in any language. The selected
items therefore share their answer letter across the six languages
(Appendix~\ref{app:prompts}).

\paragraph{TriviaQA.}
We use a 960-item six-language version of TriviaQA
\citep{joshi2017triviaqa} whose English questions, canonical answers
and aliases match the validation source. The five other language
versions of the questions and canonical answers were translated with
Claude Opus~4.7 and checked by the authors; aliases exist only in
English. The response instruction requests
a short answer.

\subsection{Factorial Units and Counts}
For each item and model, six question languages crossed with six
requested response languages produce 36 responses. The same
response is then evaluated under six verifier languages. Thus each
model--dataset pair contains 34,560 responses and 207,360 verifier
evaluations. The ten-model, three-dataset panel comprises
1,036,800 responses and 6,220,800 verifier evaluations.

On LSQA, a content-language match additionally refers to the item's
subset;
the four MMMLU subject groups are not native languages.

\subsection{Model Panel}
The probe reads next-token logits, so the panel comprises ten
instruction-tuned open-weight models from five developers:
Qwen2.5-3B, -7B and -14B \citep{qwen25}, Qwen3-8B \citep{qwen3},
Llama-3.1-8B and Llama-3.2-3B \citep{llama3,llama32}, Gemma-2-9B and
Gemma-3-12B \citep{gemma2,gemma3}, Mistral-7B-v0.3 \citep{mistral}
and the multilingual Aya Expanse 8B \citep{dang2024ayaexpanse}.

\subsection{Generation and Runtime}
\label{sec:generation}
The generation manifests record greedy decoding
(\texttt{do\_sample=False}), seed 3407, a maximum of 256 new tokens,
and bfloat16 inference. Generation and verification use a
2,048-token input limit, with batch sizes 2, 4, 8, or 16 according
to the slice. The recorded environment is Python 3.10.19,
PyTorch 2.9.0+cu128, and Transformers 4.57.3 on RTX A6000 or
RTX 3090 GPUs.

Qwen3-8B generates with its chat template's thinking mode disabled
(\texttt{enable\_\allowbreak thinking=False}, recorded in the generation
manifests), as in its verification (Section~\ref{sec:confidence}), so
its responses contain no reasoning block. In the template's default
thinking mode, two-thirds of its responses had ended inside an
unfinished reasoning block within the 256-token limit, so we
regenerated them in this mode after the human review
(Section~\ref{sec:human}); no other model, prompt or setting changed.

Inputs beyond the 2,048-token limit are truncated from the right.
This affects 0.02\%--0.03\% of prompts, all from 16 long MMMLU items,
and excluding the 357 affected responses changes no MMMLU result
(Appendix~\ref{app:robustness}).

LSQA and TriviaQA place a localized response-language directive
and the question together in one user message. MMMLU instead
uses an English system directive requesting an initial option
letter and a brief justification in the requested language, with
the localized question and choices in the user message.
Appendix~\ref{app:prompts} gives the English text and the associated
machine-readable specification.

\subsection{Confidence Extraction}
\label{sec:confidence}
The responding model also supplies the verifier logits. The
verifier instruction is localized into $V$, while the inserted
question remains in $Q$ and the generated response is unchanged.
Let $T$ and $F$ denote the deduplicated first-token IDs of the
English True and False surface variants. Adapting the $\ptrue$
protocol of \citet{kadavath2022language}, the implemented score is
\begin{equation}
\ptrue = \sigma\!\left(
\max_{t\in T}\ell_t-\max_{f\in F}\ell_f\right).
\end{equation}
The variants are True/true and False/false, each with and without
a leading space. This is maximum-logit aggregation, not a sum
of full-string probabilities. The target surface forms remain
English across all verifier languages, so that $V$ changes the
instruction but not the scored tokens. ``True'' and ``False'' are
also single tokens in all ten tokenizers, whereas for French,
Japanese, Korean and Hindi at least one word of the translated pair
spans several tokens in at least eight of them, so a first-token
probability would score only its beginning
(Appendix~\ref{app:prompts}).

Main-panel Qwen3 verification sets
\texttt{enable\_\allowbreak thinking=False}. A separate LSQA run
reads the same responses' verifier logits with thinking enabled, as a
template-mode ablation that generates no reasoning trace (Appendix~\ref{app:thinking}).

\section{Correctness Labels and Validation}
\label{sec:labels}
\subsection{Correctness Sources}
LSQA and TriviaQA responses are free text and are labeled by the
automated judge described below. MMMLU responses are labeled by
exact match between the extracted option and the gold option. We
adopted this after the human validation in Section~\ref{sec:human}
showed that the judge over-credits MMMLU answers; the judge's MMMLU
labels are retained only as a sensitivity analysis.

The extractor first applies the two regular
expressions used at generation time: a leading option letter followed
by punctuation, or an explicit ``X)'', ``X.'' or ``X:'' marker. Only
when neither matches does it accept a lone option letter, optionally
wrapped in Markdown emphasis, that forms the entire first line (e.g.\
\texttt{C} or \texttt{**D**}). A response without an extractable
final option is scored incorrect. Across the 345,600 MMMLU responses,
325,369 (94.1\%) are matched by the generation-time expressions,
14,994 (4.3\%) by the first-line rule and 5,237 (1.5\%) have no
extractable option. That share is at most 4.2\% per model and ranges
only from 0.9\% to 2.3\% across question languages and from 1.1\% to
2.1\% across requested response languages.

\subsection{Automated Judge}
The correctness judge is
\texttt{Unbabel/\allowbreak M-Prometheus-14B} \citep{pombal2025mprometheus},
an open multilingual LLM judge scoring each
response 1--5. It is fine-tuned from Qwen2.5-14B-Instruct,
one of the evaluated models; on the human sample it under-credits that
model's free-text answers more than any other model's
(Appendix~\ref{app:robustness}).
Parse-valid scores 4--5 map to correct and scores 1--3 to
incorrect. Malformed outputs are kept in the raw judge records and
counted as incorrect when the binary labels are derived; this
conservative fallback is a policy decision, not verified
incorrectness.

On a 1,440-response MMMLU pilot it reached a balanced accuracy of
77.4\%, above the 72.1\% of M-Prometheus-7B; Section~\ref{sec:human}
validates it against human labels.

The judge receives the question and any choices in $Q$, the
unchanged response, the response-language reference answer,
an English reference cross-check, and supplied English aliases.
Its rubric evaluates central factual correctness and explicitly
does not penalize code-switching or a different response language
when the core fact is correct. Language compliance therefore
requires a separate label (Appendix~\ref{app:prompts} gives the prompt
and parser).

The judge uses greedy decoding in bfloat16, no judge prompt is
truncated, and every response received a judge output.
Score-3 and malformed outputs, both counted incorrect, make up 6.3\%
of LSQA and 1.9\% of TriviaQA judgments and are most frequent for
requested Hindi responses (Appendix~\ref{app:robustness});
Section~\ref{sec:robustness} reports results without them.

\subsection{Human Validation}
\label{sec:human}
To test the judge on our responses, we froze a review sample
before judging finished and without using judge scores: one response
from each dataset--model--$Q$--$R$ cell
($3\times10\times6\times6=1{,}080$), so every eligible response had
inclusion probability $1/960$. The five authors, one per requested
response language (one author covered two), rated the 180 responses of
each language for factual correctness and, separately, for
response-language compliance, and audited 72 translation/reference
units per language. All reported understanding the original
questions in all six languages, aided by parallel versions in their
own language and in English, and used no LLM assistance; model
identities, judge outputs and confidence scores were hidden. Primary
ratings were locked before any
automated label was joined. Qwen3-8B's responses were regenerated
after the review (Section~\ref{sec:generation}), so its 108 ratings
describe superseded responses and are excluded; of the remaining 972
(162 per language), 37 were marked unjudgeable. Table~\ref{tab:human}
reports agreement with 95\% intervals from a bootstrap over base
items (2,000 replicates).

Overall, the judge agrees with the human label on 80.3\% of
responses (balanced accuracy 0.806; Cohen's $\kappa=0.60$, reported
descriptively). The disagreement is not symmetric across datasets.
On LSQA and TriviaQA the judge is conservative: its correct rate is
5.7 and 7.7 points below the human rate. On LSQA this gap ranges by
requested language from 0.0 points (English, Korean) to 11.3
(Chinese) and 14.0 (Hindi), with about 50 responses each, and on both
benchmarks agreement is higher when question and response share a
language (0.915 against 0.845; Appendix~\ref{app:robustness}).
Contrasts between responses ($R$, $Q{=}R$) can therefore absorb
scorer error, whereas verifier-language contrasts ($V$, $R{=}V$,
$Q{=}V$) compare identical responses and labels. On MMMLU it over-credits:
it labels 66.0\% of responses correct against 37.9\% for the
reviewers (specificity 0.526), and it accepts 55 responses whose
selected option the reviewer marked wrong. Exact option match
agrees with the reviewer's option-level judgment on 98.1\% of MMMLU
responses and with the reviewer's overall correctness on 85.6\%;
the remaining gap mainly reflects correct options with incorrect
justifications, which the reviewers counted as incorrect. On the
sample, all 14 options recovered by the first-line rule match the
reviewer's selected option. The extraction rules were fixed after the
ratings were locked, so these agreement rates are in-sample.

The conservative fallbacks behave differently. Of the 41 judgeable
responses whose judge output was malformed, reviewers marked 32
incorrect; of the 13 with judge score 3, they marked 10 correct and
3 incorrect.

\begin{table}[t]
\centering
\small
\setlength{\tabcolsep}{3pt}
\begin{tabular}{llccc}
\toprule
Scorer & Subset & $n$ & Accuracy & $\Delta$ \\
\midrule\midrule
Judge & All & 935 & .803 [.777,.829] & $+.047$ \\
Judge & LSQA & 316 & .816 [.771,.860] & $-.057$ \\
Judge & TriviaQA & 313 & .898 [.865,.930] & $-.077$ \\
Judge & MMMLU & 306 & .693 [.638,.747] & $+.281$ \\
EM & MMMLU & 306 & .856 [.814,.896] & $+.137$ \\
EM & MMMLU$^{a}$ & 310 & .981 [.961,.997] & $+.000$ \\
\bottomrule
\end{tabular}
\caption{Agreement of automated labels with single primary human
ratings (Qwen3-8B excluded). EM: exact option match. $\Delta$: the scorer's correct rate
minus the human correct rate. $^{a}$Against the reviewer's
option-level label rather than overall correctness.}
\label{tab:human}
\end{table}

In a separate audit of 432 item--language units, reviewers rated 38
of the 144 MMMLU units as having major defects (23 whose options no
longer matched the source choices), against one each for LSQA and
TriviaQA.
Only 61.4\% of sampled responses complied with the requested response
language (Appendix~\ref{app:human}); requests for English were least
often honored (34.0\%), and Llama-3.2-3B was the least compliant model
(43.5\%). Judge--human agreement is 0.815 on compliant responses but
0.743 on noncompliant and 0.702 on mixed ones.

\section{Analysis}
\label{sec:analysis}
We distinguish accuracy from confidence discrimination (AUROC)
and absolute calibration (15 equal-width-bin ECE,
\citealp{guo2017calibration}; and the Brier score, a strictly proper
scoring rule, \citealp{gneiting2007strictly}).

For response labels $y_i\in\{0,1\}$ and confidence $p_i$, we use
$\mathrm{Acc}=N^{-1}\sum_i y_i$ and
$\mathrm{Brier}=N^{-1}\sum_i(p_i-y_i)^2$.
For the 15 equal-width confidence bins $B_b$, calibration error is
\begin{equation}
\mathrm{ECE}_{15}=\sum_{b=1}^{15}\frac{|B_b|}{N}
\left|\overline{p}_{B_b}-\overline{y}_{B_b}\right|,
\end{equation}
where empty bins contribute zero. Correctness is shared by the
six verifier-language copies of a response, which add no accuracy
observations.

For an LSQA item with content-subset language $s$, the 36 $Q/R$
conditions comprise one native match ($Q=R=s$), five question-only
matches ($Q=s,R\ne s$), five response-only matches ($R=s,Q\ne s$),
and 25 matching neither.

\paragraph{Aggregation.} A configuration cell is one
dataset, model, $Q$, $R$ and $V$; it pools its 960 item rows.
Summary tables report unweighted means of the listed cells or units,
so each model and verifier language counts equally. All 6,480
cells contain both label classes, so AUROC is defined
throughout.

\paragraph{Accuracy decomposition.} For each model, on LSQA and
MMMLU, we fit a row-level linear model of response correctness with
sum-coded $Q$, $R$ and a stratum $S$ (the LSQA content subset or the
MMMLU subject group) and all two-way interactions, using one
verifier-language copy per response. We report each Type~III sum of
squares as a share of the explained sum, which excludes the intercept
and residual, and average the shares over models. TriviaQA has no
comparable stratum.

\paragraph{Configuration-level decomposition.} For each model,
dataset and metric we fit the analogous Type~III model to
configuration cells with factors $Q$, $R$, $V$ (and the content
subset for LSQA, on subset-level cells) and their two-way
interactions. These shares are provisional descriptions rather than
identified effects.

\paragraph{Matching effects.} For AUROC and ECE separately, per
dataset, we fit a linear mixed model to the 2,160 configuration cells
with fixed indicators $Q{=}R$, $R{=}V$, $Q{=}V$ and $Q{=}R{=}V$ and a
random intercept per model, by maximum likelihood in statsmodels
\citep{seabold2010statsmodels}, and report Wald
95\% intervals. As a secondary check we resample the ten models with
replacement (200 replicates, seed 3407) and report the fraction of
replicates whose coefficient has the primary fit's sign. A
model-level test fits the same regression separately for each model
and compares the mean of the ten coefficients with a $t$ distribution
(9 df), allowing the effects to differ across models. None of these
accounts for base items shared across configuration cells,
ten models support limited model-level generalization, and we do not
adjust for the 24 coefficients reported.

\section{Results}
\subsection{Overview}
\begin{figure}[t]
\centering
\includegraphics[width=\columnwidth]{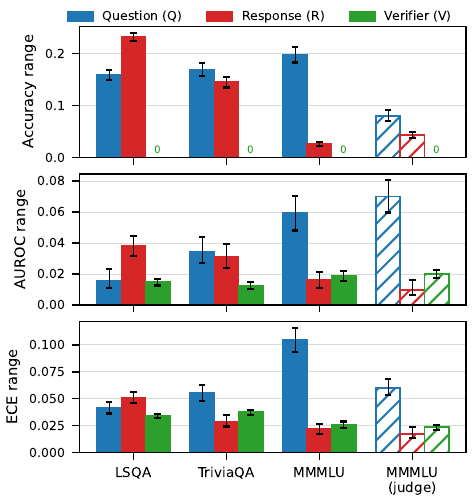}
\caption{Range (maximum minus minimum) of each axis's six
language-marginal means, with 95\% item-bootstrap intervals. $V$
cannot change accuracy. Hatched: MMMLU scored by the judge instead of
exact match. Ranges are non-negative; the intervals show sampling
variability, not tests against zero.}
\label{fig:effects}
\end{figure}

Table~\ref{tab:overview} gives the mean of each metric per dataset.
TriviaQA has the lowest accuracy
but the strongest discrimination; MMMLU, under exact match, has the
weakest discrimination and the worst calibration. Absolute
calibration is poor throughout, with ECE between 0.27 and 0.37. Figure~\ref{fig:effects} shows how far each language axis moves each metric.

\begin{table}[t]
\centering
\small
\begin{tabular}{lcccc}
\toprule
Dataset & Acc & AUROC & ECE$_{15}$ & Brier \\
\midrule\midrule
LSQA & .511 & .799 & .270 & .274 \\
MMMLU & .523 & .691 & .369 & .377 \\
TriviaQA & .316 & .822 & .287 & .287 \\
\bottomrule
\end{tabular}
\caption{Mean over the 2,160 configuration cells (10 models $\times$
216 $Q/R/V$ cells) per dataset. MMMLU uses exact match; LSQA and
TriviaQA use the judge.}
\label{tab:overview}
\end{table}

\subsection{MMMLU Accuracy under Exact Match}
\label{sec:results-mmmlu}
Pooled over the ten models, MMMLU accuracy is 0.523 under exact match
against 0.658 under the judge's labels, and exact match is lower for
every model (Table~\ref{tab:mmmlu-models} in
Appendix~\ref{app:additional}).

Under exact match, question language separates MMMLU accuracy far
more than requested response language: English questions score 0.198
higher than Hindi ones (95\% item-bootstrap interval [0.182, 0.213]),
whereas the six requested response languages span only 0.026, with
English highest (Figure~\ref{fig:effects}). The judge's labels would
change this picture: the English--Hindi question gap shrinks to 0.038
and English requests rank last. The same pattern holds within each
subject group (Table~\ref{tab:mmmlu-groups} in
Appendix~\ref{app:additional}).

Three features of the measurement bound this contrast. Localized
items carry most of the audited translation defects: reviewers found
major defects in 2 of 24 audited English items (doubtful source keys)
but in up to 10 of 24 Chinese and Hindi ones, mostly mistranslated or
corrupted options (e.g., fractions rendered as dates), so the
English--Hindi difference is best read as an upper bound on the
question-language effect. The design also limits the response-language
contrast: the requested language is named only in an English system
directive, and exact match reads the option letter that precedes any
response-language text. Finally, excluding the few responses without a
final option leaves the English--Hindi difference almost unchanged
(Section~\ref{sec:robustness}).

\subsection{Question and Response Language}
\label{sec:results-qr}
\begin{figure*}[t]
\centering
\includegraphics[width=\textwidth]{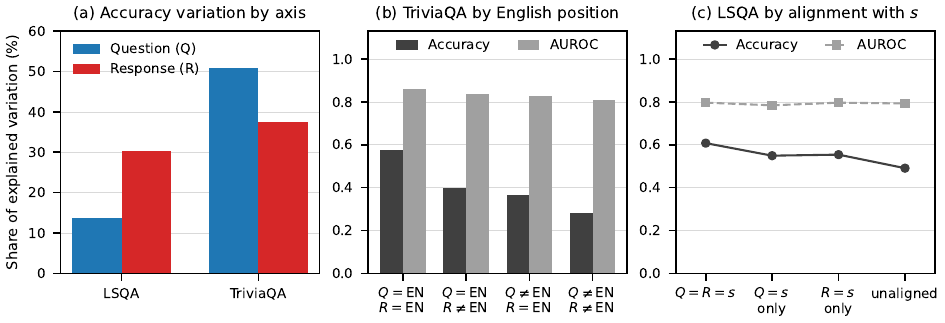}
\caption{(a) Mean share of explained accuracy variation attributed to
question and response language on LSQA and TriviaQA in the
provisional configuration-level decomposition. (b) TriviaQA accuracy and AUROC by the English position
of question and response. (c) LSQA accuracy and AUROC by alignment
with the content-subset language; ``unaligned'' pools the 25
conditions matching neither.}
\label{fig:summary}
\end{figure*}

Which language axis governs accuracy depends on the benchmark
(Figure~\ref{fig:summary}(a)). On
LSQA, the content subset and the requested response language account
for most explained accuracy variation (mean shares over models:
39.1\% and 30.4\%); on MMMLU the question language dominates
(56.6\%); for TriviaQA the provisional configuration-level
decomposition gives both languages large shares (question 51.0\%,
response 37.4\%, their interaction 11.6\%). The per-model shares
vary widely: the LSQA response-language share ranges
from 9.8\% to 55.5\%.

On TriviaQA, accuracy and AUROC are highest when both question and
response are English and lowest when neither is
(Figure~\ref{fig:summary}(b)), but calibration is worst for
non-English questions with English responses (ECE 0.303 and
Brier 0.304, against at most 0.288 and 0.291 for the other groups).
An English question raises accuracy by 0.213 when the response is
English and by 0.117 when it is not.

\begin{table}[t]
\centering
\small
\setlength{\tabcolsep}{4pt}
\begin{tabular}{lcccc}
\toprule
LSQA regime & Acc & AUROC & ECE$_{15}$ & Brier \\
\midrule\midrule
$Q{=}R{=}s$ (native) & .608 & .797 & .264 & .260 \\
$R{=}s$ only & .554 & .797 & .257 & .260 \\
$Q{=}s$ only & .549 & .784 & .261 & .265 \\
$Q{=}R{\ne}s$ & .514 & .823 & .285 & .287 \\
$Q{\ne}R$, both ${\ne}s$ & .485 & .788 & .270 & .277 \\
\bottomrule
\end{tabular}
\caption{LSQA metrics by alignment with the item's content-subset
language $s$: means over the 360 model--subset--$V$ units of each
regime, each pooling the regime's $Q/R$ conditions.}
\label{tab:lsqa-regimes}
\end{table}

On LSQA, asking and answering in the item's content language gives
the highest accuracy (Table~\ref{tab:lsqa-regimes}; Figure~\ref{fig:summary}(c)); matching only
the response or only the question recovers part of this advantage,
and about equally, so the larger share of the response language
reflects how hard it is to answer in each language on every subset
rather than alignment with the content language.
Discrimination does not follow accuracy: AUROC is highest when the
question and response share a language other than the content
language, and lowest when only the question matches; averaging
configuration cells instead of pooling them shrinks the first margin
(0.805 against 0.797; Appendix~\ref{app:robustness}). The native
advantage, native accuracy minus the mean over the 25 conditions in
which neither $Q$ nor $R$ is $s$, is large on the English and French
subsets (0.27 and 0.25, positive for all ten models), smaller on
Chinese and Japanese, and indistinguishable from zero on Korean and
Hindi. This contrast also carries each language's general strength,
since the unaligned conditions include English. With the subset and
the full $Q{\times}R$ table held fixed, asking and answering in the
content language adds 0.17 and 0.19 accuracy on the Korean and Hindi
subsets and about 0.1 on Chinese and Japanese, for all ten models
(Appendix~\ref{app:robustness}). The Korean
subset has the lowest accuracy (0.328), and its ECE exceeds the mean
of the other subsets by 0.06--0.12 in every regime.

\subsection{Language Matching and Confidence}
Matching the requested response language to the question or to the
verifier language raises AUROC on LSQA and TriviaQA by about
0.02--0.03 (Table~\ref{tab:e9}), with every model-cluster resample
and nine or ten of the ten models agreeing in sign. On MMMLU the same effects are four to six
times smaller and, although their Wald intervals exclude zero, the
model-level test does not distinguish them from zero
(Appendix~\ref{app:robustness}). The negative three-way term means
that full alignment ($Q{=}R{=}V$) adds less than the two pairwise
matches suggest, and $Q{=}V$ has no detectable effect on AUROC.

Calibration does not move with discrimination. No response-language
match lowers ECE on LSQA or TriviaQA; the positive ECE coefficients in
Table~\ref{tab:e9}, which mean worse calibration, appear in only six
to eight of the ten models, and the model-level test does not
distinguish them from zero. $Q{=}V$ lowers TriviaQA ECE in eight
models, and on MMMLU $R{=}V$ lowers ECE in nine of the ten models
although its Wald interval includes zero.

\begin{table}[t]
\centering
\footnotesize
\setlength{\tabcolsep}{1.8pt}
\begin{tabular}{llccc}
\toprule
& Term & LSQA & MMMLU & TriviaQA \\
\midrule\midrule
\multirow{4}{*}{\rotatebox{90}{AUROC}}
& $Q{=}R$ & $+.024^{*}$\,(1.00) & $+.006^{*}$\,(.940) & $+.023^{*}$\,(1.00) \\
& $R{=}V$ & $+.028^{*}$\,(1.00) & $+.005^{*}$\,(1.00) & $+.018^{*}$\,(1.00) \\
& $Q{=}V$ & $+.001$\,(.675) & $+.002$\,(.945) & $+.001$\,(.720) \\
& $Q{=}R{=}V$ & $-.027^{*}$\,(1.00) & $-.001$\,(.755) & $-.016^{*}$\,(1.00) \\
\midrule
\multirow{4}{*}{\rotatebox{90}{ECE$_{15}$}}
& $Q{=}R$ & $+.014^{*}$\,(.960) & $-.005$\,(1.00) & $+.016^{*}$\,(.990) \\
& $R{=}V$ & $-.002$\,(.625) & $-.004$\,(1.00) & $+.015^{*}$\,(.985) \\
& $Q{=}V$ & $-.004$\,(.875) & $-.007$\,(.975) & $-.008^{*}$\,(1.00) \\
& $Q{=}R{=}V$ & $-.007$\,(.745) & $+.001$\,(.810) & $-.022$\,(1.00) \\
\bottomrule
\end{tabular}
\caption{Mixed-model matching coefficients. $^{*}$Wald 95\% interval
excludes zero. Parentheses: fraction of 200 model-cluster resamples
with the same sign. Positive ECE coefficients mean worse
calibration.}
\label{tab:e9}
\end{table}

\subsection{Verifier Language}
Verifier language cannot change accuracy, but it does change
confidence quality. In the provisional decomposition,
the main effect of $V$ accounts for 6\%--18\% of explained AUROC and
ECE variation on the three benchmarks, and interactions involving $V$
add 6\%--13\%. Averaged over configurations, AUROC differs across
verifier languages by at most 0.02 on each benchmark, whereas ECE
differs by up to 0.04: French instructions give the lowest ECE on all
three benchmarks, and Korean or Chinese ones the highest. One correct
French answer receives a verifier confidence of 1.00 under English or
French instructions but below 0.001 under Chinese or Hindi ones
(Appendix~\ref{app:examples}).

\subsection{Robustness}
\label{sec:robustness}
Bracketed intervals and those in Figure~\ref{fig:effects} come from
an item bootstrap with the models held fixed, so they reflect item
sampling only. Excluding judge
fallbacks, MMMLU responses without a final option, Qwen3-8B, or
responses outside the requested language leaves the LSQA and
TriviaQA AUROC matching results and the MMMLU question-language
gap in place; keeping only responses in the requested language
removes the small MMMLU AUROC effects and shifts TriviaQA's English
advantage from the question to the response
(Appendix~\ref{app:robustness}).

\section{Conclusion}
We separated the question, requested-response and verifier languages
of a first-token $\ptrue$ probe on three benchmarks and ten models.
These languages act as experimental variables
whose effects depend on the benchmark and the metric: the question
language dominates MMMLU accuracy, the content subset and requested
response language dominate LSQA, and both matter on TriviaQA.
Matching the response language to the question or verifier language
improves discrimination on LSQA and TriviaQA without improving
calibration, and the verifier language shifts ECE by up to 0.04
without changing accuracy. Measurement choices matter too: an
unvalidated judge would have masked most of the MMMLU
question-language gap, and answers outside the requested language
appear to account for part of TriviaQA's English-question advantage.
Multilingual confidence results should therefore report all three
languages, the realized response language and a scorer validated per
benchmark. The findings concern one probe and
3B--14B open-weight models, not languages in general.

\section*{Limitations}
Our results concern one confidence estimator, a first-token $\ptrue$
probe, and ten open-weight models from 3B to 14B parameters; other
estimators and larger or closed models may behave differently. We use
one prompt template per benchmark, and model outputs are known to be
sensitive to prompt format \citep{sclar2024quantifying,zhuo2024prosa}.
Localized verifier instructions keep the English labels True and
False, which may contribute to the verifier-language effects. The
non-English MMMLU and TriviaQA items are translated from English,
which can favor English questions.

\section*{Ethics and Reproducibility}
\paragraph{Human review.} The human review was split among the five
authors, one per requested response language (one author covered two
languages), each a native or fluent speaker of the languages reviewed,
without LLM assistance (Section~\ref{sec:human}; full instructions in
Appendix~\ref{app:instructions}).
The material consisted of model outputs, items from existing
benchmarks and our LSQA items, whose celebrity questions concern
public figures; it contained no sensitive personal information. As all
annotations were provided by the authors, no external participants
were involved and no institutional ethics review was sought.

\paragraph{Risks.} The results describe one confidence probe, three
benchmarks and ten 3B--14B models. Reading them as properties of
languages, or treating first-token $\ptrue$ as a reliability guarantee
for speakers of the languages studied, could misdirect deployment
decisions. The released model outputs contain factual errors,
including about public figures, and are released for research use
only.

\paragraph{Artifacts and licenses.} We use all artifacts for
non-commercial research. Qwen2.5-7B/14B-Instruct, Qwen3-8B and
Mistral-7B-Instruct-v0.3 are released under Apache~2.0;
Qwen2.5-3B-Instruct under the Qwen Research License, which permits
research and evaluation only; Llama~3.1 and~3.2 under their community
licenses and acceptable use policies; Gemma~2 and~3 under the Gemma
Terms of Use; and Aya Expanse under CC BY-NC~4.0 with Cohere's
acceptable use policy. The model card of the M-Prometheus-14B judge
gives its license only as ``other''. MMMLU and MMLU are MIT-licensed;
TriviaQA is distributed by its authors under Apache~2.0, although they
do not own the copyright of the questions. LSQA follows the design of
MlingConf, whose release states no license, without reusing its items.
All model responses are machine-generated, and any release of them or
of LSQA must respect the non-commercial terms above.

\paragraph{Use of AI assistants.} LSQA items were generated and
translated with Claude Opus~4.7 and checked by the authors, and the
non-English TriviaQA versions were translated with the same model and
checked by the authors (Section~\ref{sec:benchmarks}). We also used an AI assistant (Claude)
to write analysis code and to draft and edit parts of this manuscript;
every reported number comes from the recorded analysis code.

\paragraph{Compute.} The final runs used about 1,717 GPU-hours on
single RTX A6000 or RTX 3090 GPUs: 359 for generation, 108 for
verification, 1,248 for the correctness judge and 2 for the
thinking-template ablation. Failed, cancelled or superseded attempts
and auxiliary runs used another 304 GPU-hours, including 216 for the
superseded thinking-mode Qwen3-8B runs.

\paragraph{Reproducibility.} The supplementary software archive
contains the generation, verification, judging and analysis code. The
data archive contains the items, the six-language prompt templates,
all 1,036,800 responses with their labels, realized-language labels
and six $\ptrue$ values, the per-cell metrics and statistics, and the
anonymized human ratings; recomputing the tables from these files
reproduces the reported values. Appendix~\ref{app:prompts} gives the
English prompts and the judge prompt. The two archives also record the
selected item identifiers, the model revisions and runtime versions,
and the SHA-256 hashes of the inputs, analysis code and outputs behind
every reported number.

\bibliography{custom}

@article{kadavath2022language,
  title   = {Language Models (Mostly) Know What They Know},
  author  = {Kadavath, Saurav and
             Conerly, Tom and
             Askell, Amanda and
             Henighan, Tom and
             Drain, Dawn and
             Perez, Ethan and
             Schiefer, Nicholas and
             Hatfield-Dodds, Zac and
             DasSarma, Nova and
             Tran-Johnson, Eli and
             Johnston, Scott and
             El-Showk, Sheer and
             Jones, Andy and
             Elhage, Nelson and
             Hume, Tristan and
             Chen, Anna and
             Bai, Yuntao and
             Bowman, Sam and
             Fort, Stanislav and
             Ganguli, Deep and
             Hernandez, Danny and
             Jacobson, Josh and
             Kernion, Jackson and
             Kravec, Shauna and
             Lovitt, Liane and
             Ndousse, Kamal and
             Olsson, Catherine and
             Ringer, Sam and
             Amodei, Dario and
             Brown, Tom and
             Clark, Jack and
             Joseph, Nicholas and
             Mann, Ben and
             McCandlish, Sam and
             Olah, Chris and
             Kaplan, Jared},
  journal = {arXiv preprint arXiv:2207.05221},
  year    = {2022},
  url     = {https://arxiv.org/abs/2207.05221}
}

@inproceedings{xue2024mlingconf,
  title     = {{MlingConf}: A Comprehensive Study of Multilingual Confidence Estimation on Large Language Models},
  author    = {Xue, Boyang and Wang, Hongru and Wang, Rui and Wang, Sheng and Wang, Zezhong and Du, Yiming and Liang, Bin and Zhang, Wenxuan and Wong, Kam-Fai},
  booktitle = {Findings of the Association for Computational Linguistics: ACL 2025},
  pages     = {2535--2556},
  year      = {2025},
  address   = {Vienna, Austria},
  publisher = {Association for Computational Linguistics},
  doi       = {10.18653/v1/2025.findings-acl.129},
  url       = {https://aclanthology.org/2025.findings-acl.129/}
}

@article{yang2024can,
  title   = {On Calibration of Multilingual Question Answering {LLMs}},
  author  = {Yang, Yahan and Dan, Soham and Roth, Dan and Lee, Insup},
  journal = {Transactions on Machine Learning Research},
  issn    = {2835-8856},
  year    = {2026},
  url     = {https://openreview.net/forum?id=4klghu2PTj}
}

@article{zhou2025layerwise,
  title   = {Beyond the Final Layer: Intermediate Representations for Better Multilingual Calibration in Large Language Models},
  author  = {Zhou, Ej and Zhang, Caiqi and Hu, Tiancheng and Li, Chengzu and Collier, Nigel and Vuli{\'c}, Ivan and Korhonen, Anna},
  journal = {arXiv preprint arXiv:2510.03136},
  year    = {2025},
  url     = {https://arxiv.org/abs/2510.03136}
}

@inproceedings{yoon2025reasoning,
  title     = {Reasoning Models Better Express Their Confidence},
  author    = {Yoon, Dongkeun and Kim, Seungone and Yang, Sohee and Kim, Sunkyoung and Kim, Soyeon and Kim, Yongil and Choi, Eunbi and Kim, Yireun and Seo, Minjoon},
  booktitle = {Advances in Neural Information Processing Systems},
  volume    = {38},
  pages     = {103869--103896},
  year      = {2025},
  publisher = {Curran Associates, Inc.},
  doi       = {10.52202/085713-3469},
  url       = {https://proceedings.neurips.cc/paper_files/paper/2025/hash/95ee7181fe256439dca74ec794b33bb6-Abstract-Conference.html}
}

@inproceedings{tian2023just,
  title     = {Just Ask for Calibration: Strategies for Eliciting Calibrated Confidence Scores from Language Models Fine-Tuned with Human Feedback},
  author    = {Tian, Katherine and Mitchell, Eric and Zhou, Allan and Sharma, Archit and Rafailov, Rafael and Yao, Huaxiu and Finn, Chelsea and Manning, Christopher D.},
  booktitle = {Proceedings of the 2023 Conference on Empirical Methods in Natural Language Processing},
  pages     = {5433--5442},
  year      = {2023},
  address   = {Singapore},
  publisher = {Association for Computational Linguistics},
  doi       = {10.18653/v1/2023.emnlp-main.330},
  url       = {https://aclanthology.org/2023.emnlp-main.330/}
}

@inproceedings{geng2024survey,
  title     = {A Survey of Confidence Estimation and Calibration in Large Language Models},
  author    = {Geng, Jiahui and Cai, Fengyu and Wang, Yuxia and Koeppl, Heinz and Nakov, Preslav and Gurevych, Iryna},
  booktitle = {Proceedings of the 2024 Conference of the North American Chapter of the Association for Computational Linguistics: Human Language Technologies (Volume 1: Long Papers)},
  pages     = {6577--6595},
  year      = {2024},
  address   = {Mexico City, Mexico},
  publisher = {Association for Computational Linguistics},
  doi       = {10.18653/v1/2024.naacl-long.366},
  url       = {https://aclanthology.org/2024.naacl-long.366/}
}

@inproceedings{xiong2024can,
  title     = {Can {LLMs} Express Their Uncertainty? {A}n Empirical Evaluation of Confidence Elicitation in {LLMs}},
  author    = {Xiong, Miao and Hu, Zhiyuan and Lu, Xinyang and Li, Yifei and Fu, Jie and He, Junxian and Hooi, Bryan},
  booktitle = {The Twelfth International Conference on Learning Representations (ICLR)},
  year      = {2024},
  url       = {https://openreview.net/forum?id=gjeQKFxFpZ}
}

@inproceedings{wendler2024llamas,
  title     = {Do Llamas Work in {E}nglish? {O}n the Latent Language of Multilingual Transformers},
  author    = {Wendler, Chris and Veselovsky, Veniamin and Monea, Giovanni and West, Robert},
  booktitle = {Proceedings of the 62nd Annual Meeting of the Association for Computational Linguistics (Volume 1: Long Papers)},
  pages     = {15366--15394},
  year      = {2024},
  address   = {Bangkok, Thailand},
  publisher = {Association for Computational Linguistics},
  doi       = {10.18653/v1/2024.acl-long.820},
  url       = {https://aclanthology.org/2024.acl-long.820/}
}

@inproceedings{qi2025crosslingual,
  title     = {When Models Reason in Your Language: Controlling Thinking Language Comes at the Cost of Accuracy},
  author    = {Qi, Jirui and Chen, Shan and Xiong, Zidi and Fern{\'a}ndez, Raquel and Bitterman, Danielle S. and Bisazza, Arianna},
  booktitle = {Findings of the Association for Computational Linguistics: EMNLP 2025},
  pages     = {20279--20296},
  year      = {2025},
  address   = {Suzhou, China},
  publisher = {Association for Computational Linguistics},
  doi       = {10.18653/v1/2025.findings-emnlp.1103},
  url       = {https://aclanthology.org/2025.findings-emnlp.1103/}
}

@inproceedings{shi2023language,
  title     = {Language Models are Multilingual Chain-of-Thought Reasoners},
  author    = {Shi, Freda and Suzgun, Mirac and Freitag, Markus and Wang, Xuezhi and Srivats, Suraj and Vosoughi, Soroush and Chung, Hyung Won and Tay, Yi and Ruder, Sebastian and Zhou, Denny and Das, Dipanjan and Wei, Jason},
  booktitle = {The Eleventh International Conference on Learning Representations (ICLR)},
  year      = {2023},
  url       = {https://openreview.net/forum?id=fR3wGCk-IXp}
}

@misc{openai2024mmmlu,
  title  = {Multilingual Massive Multitask Language Understanding ({MMMLU})},
  author = {{OpenAI}},
  year   = {2024},
  note   = {Dataset available at \url{https://huggingface.co/datasets/openai/MMMLU}},
  url    = {https://huggingface.co/datasets/openai/MMMLU}
}

@article{qwen25,
  title   = {{Qwen2.5} Technical Report},
  author  = {{Qwen Team}},
  journal = {arXiv preprint arXiv:2412.15115},
  year    = {2024},
  url     = {https://arxiv.org/abs/2412.15115}
}

@article{llama3,
  title   = {The {Llama} 3 Herd of Models},
  author  = {Grattafiori, Aaron and
             Dubey, Abhimanyu and
             Jauhri, Abhinav and
             Pandey, Abhinav and
             Kadian, Abhishek and
             Al-Dahle, Ahmad and
             Letman, Aiesha and
             Mathur, Akhil and
             Schelten, Alan and
             Vaughan, Alex and
             Yang, Amy and
             Fan, Angela and
             Goyal, Anirudh and
             Hartshorn, Anthony and
             Yang, Aobo and
             Mitra, Archi and
             Sravankumar, Archie and
             Korenev, Artem and
             Hinsvark, Arthur and
             Rao, Arun and
             Zhang, Aston and
             Rodriguez, Aurelien and
             Gregerson, Austen and
             Spataru, Ava and
             Roziere, Baptiste and
             Biron, Bethany and
             Tang, Binh and
             Chern, Bobbie and
             Caucheteux, Charlotte and
             Nayak, Chaya and
             Bi, Chloe and
             Marra, Chris and
             McConnell, Chris and
             Keller, Christian and
             Touret, Christophe and
             Wu, Chunyang and
             Wong, Corinne and
             Ferrer, Cristian Canton and
             Nikolaidis, Cyrus and
             Allonsius, Damien and
             Song, Daniel and
             Pintz, Danielle and
             Livshits, Danny and
             Wyatt, Danny and
             Esiobu, David and
             Choudhary, Dhruv and
             Mahajan, Dhruv and
             Garcia-Olano, Diego and
             Perino, Diego and
             Hupkes, Dieuwke and
             Lakomkin, Egor and
             AlBadawy, Ehab and
             Lobanova, Elina and
             Dinan, Emily and
             Smith, Eric Michael and
             Radenovic, Filip and
             Guzm{\'a}n, Francisco and
             Zhang, Frank and
             Synnaeve, Gabriel and
             Lee, Gabrielle and
             Anderson, Georgia Lewis and
             Thattai, Govind and
             Nail, Graeme and
             Mialon, Gregoire and
             Pang, Guan and
             Cucurell, Guillem and
             Nguyen, Hailey and
             Korevaar, Hannah and
             Xu, Hu and
             Touvron, Hugo and
             Zarov, Iliyan and
             Ibarra, Imanol Arrieta and
             Kloumann, Isabel and
             Misra, Ishan and
             Evtimov, Ivan and
             Zhang, Jack and
             Copet, Jade and
             Lee, Jaewon and
             Geffert, Jan and
             Vranes, Jana and
             Park, Jason and
             Mahadeokar, Jay and
             Shah, Jeet and
             van der Linde, Jelmer and
             Billock, Jennifer and
             Hong, Jenny and
             Lee, Jenya and
             Fu, Jeremy and
             Chi, Jianfeng and
             Huang, Jianyu and
             Liu, Jiawen and
             Wang, Jie and
             Yu, Jiecao and
             Bitton, Joanna and
             Spisak, Joe and
             Park, Jongsoo and
             Rocca, Joseph and
             Johnstun, Joshua and
             Saxe, Joshua and
             Jia, Junteng and
             Alwala, Kalyan Vasuden and
             Prasad, Karthik and
             Upasani, Kartikeya and
             Plawiak, Kate and
             Li, Ke and
             Heafield, Kenneth and
             Stone, Kevin and
             El-Arini, Khalid and
             Iyer, Krithika and
             Malik, Kshitiz and
             Chiu, Kuenley and
             Bhalla, Kunal and
             Lakhotia, Kushal and
             Rantala-Yeary, Lauren and
             van der Maaten, Laurens and
             Chen, Lawrence and
             Tan, Liang and
             Jenkins, Liz and
             Martin, Louis and
             Madaan, Lovish and
             Malo, Lubo and
             Blecher, Lukas and
             Landzaat, Lukas and
             de Oliveira, Luke and
             Muzzi, Madeline and
             Pasupuleti, Mahesh and
             Singh, Mannat and
             Paluri, Manohar and
             Kardas, Marcin and
             Tsimpoukelli, Maria and
             Oldham, Mathew and
             Rita, Mathieu and
             Pavlova, Maya and
             Kambadur, Melanie and
             Lewis, Mike and
             Si, Min and
             Singh, Mitesh Kumar and
             Hassan, Mona and
             Goyal, Naman and
             Torabi, Narjes and
             Bashlykov, Nikolay and
             Bogoychev, Nikolay and
             Chatterji, Niladri and
             Zhang, Ning and
             Duchenne, Olivier and
             {\c{C}}elebi, Onur and
             Alrassy, Patrick and
             Zhang, Pengchuan and
             Li, Pengwei and
             Vasic, Petar and
             Weng, Peter and
             Bhargava, Prajjwal and
             Dubal, Pratik and
             Krishnan, Praveen and
             Koura, Punit Singh and
             Xu, Puxin and
             He, Qing and
             Dong, Qingxiao and
             Srinivasan, Ragavan and
             Ganapathy, Raj and
             Calderer, Ramon and
             Cabral, Ricardo Silveira and
             Stojnic, Robert and
             Raileanu, Roberta and
             Maheswari, Rohan and
             Girdhar, Rohit and
             Patel, Rohit and
             Sauvestre, Romain and
             Polidoro, Ronnie and
             Sumbaly, Roshan and
             Taylor, Ross and
             Silva, Ruan and
             Hou, Rui and
             Wang, Rui and
             Hosseini, Saghar and
             Chennabasappa, Sahana and
             Singh, Sanjay and
             Bell, Sean and
             Kim, Seohyun Sonia and
             Edunov, Sergey and
             Nie, Shaoliang and
             Narang, Sharan and
             Raparthy, Sharath and
             Shen, Sheng and
             Wan, Shengye and
             Bhosale, Shruti and
             Zhang, Shun and
             Vandenhende, Simon and
             Batra, Soumya and
             Whitman, Spencer and
             Sootla, Sten and
             Collot, Stephane and
             Gururangan, Suchin and
             Borodinsky, Sydney and
             Herman, Tamar and
             Fowler, Tara and
             Sheasha, Tarek and
             Georgiou, Thomas and
             Scialom, Thomas and
             Speckbacher, Tobias and
             Mihaylov, Todor and
             Xiao, Tong and
             Karn, Ujjwal and
             Goswami, Vedanuj and
             Gupta, Vibhor and
             Ramanathan, Vignesh and
             Kerkez, Viktor and
             Gonguet, Vincent and
             Do, Virginie and
             Vogeti, Vish and
             Albiero, V{\'\i}tor and
             Petrovic, Vladan and
             Chu, Weiwei and
             Xiong, Wenhan and
             Fu, Wenyin and
             Meers, Whitney and
             Martinet, Xavier and
             Wang, Xiaodong and
             Wang, Xiaofang and
             Tan, Xiaoqing Ellen and
             Xia, Xide and
             Xie, Xinfeng and
             Jia, Xuchao and
             Wang, Xuewei and
             Goldschlag, Yaelle and
             Gaur, Yashesh and
             Babaei, Yasmine and
             Wen, Yi and
             Song, Yiwen and
             Zhang, Yuchen and
             Li, Yue and
             Mao, Yuning and
             Coudert, Zacharie Delpierre and
             Yan, Zheng and
             Chen, Zhengxing and
             Papakipos, Zoe and
             Singh, Aaditya and
             Srivastava, Aayushi and
             Jain, Abha and
             Kelsey, Adam and
             Shajnfeld, Adam and
             Gangidi, Adithya and
             Victoria, Adolfo and
             Goldstand, Ahuva and
             Menon, Ajay and
             Sharma, Ajay and
             Boesenberg, Alex and
             Baevski, Alexei and
             Feinstein, Allie and
             Kallet, Amanda and
             Sangani, Amit and
             Teo, Amos and
             Yunus, Anam and
             Lupu, Andrei and
             Alvarado, Andres and
             Caples, Andrew and
             Gu, Andrew and
             Ho, Andrew and
             Poulton, Andrew and
             Ryan, Andrew and
             Ramchandani, Ankit and
             Dong, Annie and
             Franco, Annie and
             Goyal, Anuj and
             Saraf, Aparajita and
             Chowdhury, Arkabandhu and
             Gabriel, Ashley and
             Bharambe, Ashwin and
             Eisenman, Assaf and
             Yazdan, Azadeh and
             James, Beau and
             Maurer, Ben and
             Leonhardi, Benjamin and
             Huang, Bernie and
             Loyd, Beth and
             De Paola, Beto and
             Paranjape, Bhargavi and
             Liu, Bing and
             Wu, Bo and
             Ni, Boyu and
             Hancock, Braden and
             Wasti, Bram and
             Spence, Brandon and
             Stojkovic, Brani and
             Gamido, Brian and
             Montalvo, Britt and
             Parker, Carl and
             Burton, Carly and
             Mejia, Catalina and
             Liu, Ce and
             Wang, Changhan and
             Kim, Changkyu and
             Zhou, Chao and
             Hu, Chester and
             Chu, Ching-Hsiang and
             Cai, Chris and
             Tindal, Chris and
             Feichtenhofer, Christoph and
             Gao, Cynthia and
             Civin, Damon and
             Beaty, Dana and
             Kreymer, Daniel and
             Li, Daniel and
             Adkins, David and
             Xu, David and
             Testuggine, Davide and
             David, Delia and
             Parikh, Devi and
             Liskovich, Diana and
             Foss, Didem and
             Wang, Dingkang and
             Le, Duc and
             Holland, Dustin and
             Dowling, Edward and
             Jamil, Eissa and
             Montgomery, Elaine and
             Presani, Eleonora and
             Hahn, Emily and
             Wood, Emily and
             Le, Eric-Tuan and
             Brinkman, Erik and
             Arcaute, Esteban and
             Dunbar, Evan and
             Smothers, Evan and
             Sun, Fei and
             Kreuk, Felix and
             Tian, Feng and
             Kokkinos, Filippos and
             Ozgenel, Firat and
             Caggioni, Francesco and
             Kanayet, Frank and
             Seide, Frank and
             Florez, Gabriela Medina and
             Schwarz, Gabriella and
             Badeer, Gada and
             Swee, Georgia and
             Halpern, Gil and
             Herman, Grant and
             Sizov, Grigory and
             Zhang, Guangyi (Jack) and
             Lakshminarayanan, Guna and
             Inan, Hakan and
             Shojanazeri, Hamid and
             Zou, Han and
             Wang, Hannah and
             Zha, Hanwen and
             Habeeb, Haroun and
             Rudolph, Harrison and
             Suk, Helen and
             Aspegren, Henry and
             Goldman, Hunter and
             Zhan, Hongyuan and
             Damlaj, Ibrahim and
             Molybog, Igor and
             Tufanov, Igor and
             Leontiadis, Ilias and
             Veliche, Irina-Elena and
             Gat, Itai and
             Weissman, Jake and
             Geboski, James and
             Kohli, James and
             Lam, Janice and
             Asher, Japhet and
             Gaya, Jean-Baptiste and
             Marcus, Jeff and
             Tang, Jeff and
             Chan, Jennifer and
             Zhen, Jenny and
             Reizenstein, Jeremy and
             Teboul, Jeremy and
             Zhong, Jessica and
             Jin, Jian and
             Yang, Jingyi and
             Cummings, Joe and
             Carvill, Jon and
             Shepard, Jon and
             McPhie, Jonathan and
             Torres, Jonathan and
             Ginsburg, Josh and
             Wang, Junjie and
             Wu, Kai and
             U, Kam Hou and
             Saxena, Karan and
             Khandelwal, Kartikay and
             Zand, Katayoun and
             Matosich, Kathy and
             Veeraraghavan, Kaushik and
             Michelena, Kelly and
             Li, Keqian and
             Jagadeesh, Kiran and
             Huang, Kun and
             Chawla, Kunal and
             Huang, Kyle and
             Chen, Lailin and
             Garg, Lakshya and
             A, Lavender and
             Silva, Leandro and
             Bell, Lee and
             Zhang, Lei and
             Guo, Liangpeng and
             Yu, Licheng and
             Moshkovich, Liron and
             Wehrstedt, Luca and
             Khabsa, Madian and
             Avalani, Manav and
             Bhatt, Manish and
             Mankus, Martynas and
             Hasson, Matan and
             Lennie, Matthew and
             Reso, Matthias and
             Groshev, Maxim and
             Naumov, Maxim and
             Lathi, Maya and
             Keneally, Meghan and
             Liu, Miao and
             Seltzer, Michael L. and
             Valko, Michal and
             Restrepo, Michelle and
             Patel, Mihir and
             Vyatskov, Mik and
             Samvelyan, Mikayel and
             Clark, Mike and
             Macey, Mike and
             Wang, Mike and
             Hermoso, Miquel Jubert and
             Metanat, Mo and
             Rastegari, Mohammad and
             Bansal, Munish and
             Santhanam, Nandhini and
             Parks, Natascha and
             White, Natasha and
             Bawa, Navyata and
             Singhal, Nayan and
             Egebo, Nick and
             Usunier, Nicolas and
             Mehta, Nikhil and
             Laptev, Nikolay Pavlovich and
             Dong, Ning and
             Cheng, Norman and
             Chernoguz, Oleg and
             Hart, Olivia and
             Salpekar, Omkar and
             Kalinli, Ozlem and
             Kent, Parkin and
             Parekh, Parth and
             Saab, Paul and
             Balaji, Pavan and
             Rittner, Pedro and
             Bontrager, Philip and
             Roux, Pierre and
             Dollar, Piotr and
             Zvyagina, Polina and
             Ratanchandani, Prashant and
             Yuvraj, Pritish and
             Liang, Qian and
             Alao, Rachad and
             Rodriguez, Rachel and
             Ayub, Rafi and
             Murthy, Raghotham and
             Nayani, Raghu and
             Mitra, Rahul and
             Parthasarathy, Rangaprabhu and
             Li, Raymond and
             Hogan, Rebekkah and
             Battey, Robin and
             Wang, Rocky and
             Howes, Russ and
             Rinott, Ruty and
             Mehta, Sachin and
             Siby, Sachin and
             Bondu, Sai Jayesh and
             Datta, Samyak and
             Chugh, Sara and
             Hunt, Sara and
             Dhillon, Sargun and
             Sidorov, Sasha and
             Pan, Satadru and
             Mahajan, Saurabh and
             Verma, Saurabh and
             Yamamoto, Seiji and
             Ramaswamy, Sharadh and
             Lindsay, Shaun and
             Lindsay, Shaun and
             Feng, Sheng and
             Lin, Shenghao and
             Zha, Shengxin Cindy and
             Patil, Shishir and
             Shankar, Shiva and
             Zhang, Shuqiang and
             Zhang, Shuqiang and
             Wang, Sinong and
             Agarwal, Sneha and
             Sajuyigbe, Soji and
             Chintala, Soumith and
             Max, Stephanie and
             Chen, Stephen and
             Kehoe, Steve and
             Satterfield, Steve and
             Govindaprasad, Sudarshan and
             Gupta, Sumit and
             Deng, Summer and
             Cho, Sungmin and
             Virk, Sunny and
             Subramanian, Suraj and
             Choudhury, Sy and
             Goldman, Sydney and
             Remez, Tal and
             Glaser, Tamar and
             Best, Tamara and
             Koehler, Thilo and
             Robinson, Thomas and
             Li, Tianhe and
             Zhang, Tianjun and
             Matthews, Tim and
             Chou, Timothy and
             Shaked, Tzook and
             Vontimitta, Varun and
             Ajayi, Victoria and
             Montanez, Victoria and
             Mohan, Vijai and
             Kumar, Vinay Satish and
             Mangla, Vishal and
             Ionescu, Vlad and
             Poenaru, Vlad and
             Mihailescu, Vlad Tiberiu and
             Ivanov, Vladimir and
             Li, Wei and
             Wang, Wenchen and
             Jiang, Wenwen and
             Bouaziz, Wes and
             Constable, Will and
             Tang, Xiaocheng and
             Wu, Xiaojian and
             Wang, Xiaolan and
             Wu, Xilun and
             Gao, Xinbo and
             Kleinman, Yaniv and
             Chen, Yanjun and
             Hu, Ye and
             Jia, Ye and
             Qi, Ye and
             Li, Yenda and
             Zhang, Yilin and
             Zhang, Ying and
             Adi, Yossi and
             Nam, Youngjin and
             Wang, Yu (Sid) and
             Zhao, Yu and
             Hao, Yuchen and
             Qian, Yundi and
             Li, Yunlu and
             He, Yuzi and
             Rait, Zach and
             DeVito, Zachary and
             Rosnbrick, Zef and
             Wen, Zhaoduo and
             Yang, Zhenyu and
             Zhao, Zhiwei and
             Ma, Zhiyu},
  journal = {arXiv preprint arXiv:2407.21783},
  year    = {2024},
  url     = {https://arxiv.org/abs/2407.21783}
}

@inproceedings{zheng2024large,
  title     = {Large Language Models Are Not Robust Multiple Choice Selectors},
  author    = {Zheng, Chujie and Zhou, Hao and Meng, Fandong and Zhou, Jie and Huang, Minlie},
  booktitle = {The Twelfth International Conference on Learning Representations (ICLR)},
  year      = {2024},
  url       = {https://openreview.net/forum?id=shr9PXz7T0}
}

@inproceedings{wang2024answer,
  title     = {``{M}y Answer is {C}'': First-Token Probabilities Do Not Match Text Answers in Instruction-Tuned Language Models},
  author    = {Wang, Xinpeng and Ma, Bolei and Hu, Chengzhi and Weber-Genzel, Leon and R{\"o}ttger, Paul and Kreuter, Frauke and Hovy, Dirk and Plank, Barbara},
  booktitle = {Findings of the Association for Computational Linguistics: ACL 2024},
  pages     = {7407--7416},
  year      = {2024},
  address   = {Bangkok, Thailand},
  publisher = {Association for Computational Linguistics},
  doi       = {10.18653/v1/2024.findings-acl.441},
  url       = {https://aclanthology.org/2024.findings-acl.441/}
}

@inproceedings{wang2024look,
  title     = {Look at the Text: Instruction-Tuned Language Models are More Robust Multiple Choice Selectors than You Think},
  author    = {Wang, Xinpeng and Hu, Chengzhi and Ma, Bolei and R{\"o}ttger, Paul and Plank, Barbara},
  booktitle = {First Conference on Language Modeling (COLM)},
  year      = {2024},
  url       = {https://openreview.net/forum?id=qHdSA85GyZ}
}

@inproceedings{sclar2024quantifying,
  title     = {Quantifying Language Models' Sensitivity to Spurious Features in Prompt Design or: How {I} learned to start worrying about prompt formatting},
  author    = {Sclar, Melanie and Choi, Yejin and Tsvetkov, Yulia and Suhr, Alane},
  booktitle = {The Twelfth International Conference on Learning Representations (ICLR)},
  year      = {2024},
  url       = {https://openreview.net/forum?id=RIu5lyNXjT}
}

@inproceedings{zhuo2024prosa,
  title     = {{ProSA}: Assessing and Understanding the Prompt Sensitivity of {LLMs}},
  author    = {Zhuo, Jingming and Zhang, Songyang and Fang, Xinyu and Duan, Haodong and Lin, Dahua and Chen, Kai},
  booktitle = {Findings of the Association for Computational Linguistics: EMNLP 2024},
  pages     = {1950--1976},
  year      = {2024},
  address   = {Miami, Florida, USA},
  publisher = {Association for Computational Linguistics},
  doi       = {10.18653/v1/2024.findings-emnlp.108},
  url       = {https://aclanthology.org/2024.findings-emnlp.108/}
}

@inproceedings{pezeshkpour2024large,
  title     = {Large Language Models Sensitivity to the Order of Options in Multiple-Choice Questions},
  author    = {Pezeshkpour, Pouya and Hruschka, Estevam},
  booktitle = {Findings of the Association for Computational Linguistics: NAACL 2024},
  pages     = {2006--2017},
  year      = {2024},
  address   = {Mexico City, Mexico},
  publisher = {Association for Computational Linguistics},
  doi       = {10.18653/v1/2024.findings-naacl.130},
  url       = {https://aclanthology.org/2024.findings-naacl.130/}
}

@article{lin2022teaching,
  title   = {Teaching Models to Express Their Uncertainty in Words},
  author  = {Lin, Stephanie and Hilton, Jacob and Evans, Owain},
  journal = {Transactions on Machine Learning Research},
  issn    = {2835-8856},
  year    = {2022},
  url     = {https://openreview.net/forum?id=8s8K2UZGTZ}
}

@article{farquhar2024detecting,
  title   = {Detecting hallucinations in large language models using semantic entropy},
  author  = {Farquhar, Sebastian and Kossen, Jannik and Kuhn, Lorenz and Gal, Yarin},
  journal = {Nature},
  volume  = {630},
  number  = {8017},
  pages   = {625--630},
  year    = {2024},
  doi     = {10.1038/s41586-024-07421-0},
  url     = {https://www.nature.com/articles/s41586-024-07421-0}
}

@inproceedings{qi2023cross,
  title     = {Cross-Lingual Consistency of Factual Knowledge in Multilingual Language Models},
  author    = {Qi, Jirui and Fern{\'a}ndez, Raquel and Bisazza, Arianna},
  booktitle = {Proceedings of the 2023 Conference on Empirical Methods in Natural Language Processing},
  pages     = {10650--10666},
  year      = {2023},
  address   = {Singapore},
  publisher = {Association for Computational Linguistics},
  doi       = {10.18653/v1/2023.emnlp-main.658},
  url       = {https://aclanthology.org/2023.emnlp-main.658/}
}

@inproceedings{wang2025lost,
  title     = {Lost in Multilinguality: Dissecting Cross-lingual Factual Inconsistency in Transformer Language Models},
  author    = {Wang, Mingyang and Adel, Heike and Lange, Lukas and Liu, Yihong and Nie, Ercong and Str{\"o}tgen, Jannik and Schuetze, Hinrich},
  booktitle = {Proceedings of the 63rd Annual Meeting of the Association for Computational Linguistics (Volume 1: Long Papers)},
  pages     = {5075--5094},
  year      = {2025},
  address   = {Vienna, Austria},
  publisher = {Association for Computational Linguistics},
  doi       = {10.18653/v1/2025.acl-long.253},
  url       = {https://aclanthology.org/2025.acl-long.253/}
}

@article{gemma2,
  title   = {{Gemma 2}: Improving Open Language Models at a Practical Size},
  author  = {{Gemma Team}},
  journal = {arXiv preprint arXiv:2408.00118},
  year    = {2024},
  url     = {https://arxiv.org/abs/2408.00118}
}

@article{mistral,
  title   = {{Mistral 7B}},
  author  = {Jiang, Albert Q. and Sablayrolles, Alexandre and Mensch, Arthur and Bamford, Chris and Chaplot, Devendra Singh and de las Casas, Diego and Bressand, Florian and Lengyel, Gianna and Lample, Guillaume and Saulnier, Lucile and Renard Lavaud, L{\'e}lio and Lachaux, Marie-Anne and Stock, Pierre and Le Scao, Teven and Lavril, Thibaut and Wang, Thomas and Lacroix, Timoth{\'e}e and El Sayed, William},
  journal = {arXiv preprint arXiv:2310.06825},
  year    = {2023},
  url     = {https://arxiv.org/abs/2310.06825}
}

@article{gneiting2007strictly,
  title   = {Strictly Proper Scoring Rules, Prediction, and Estimation},
  author  = {Gneiting, Tilmann and Raftery, Adrian E.},
  journal = {Journal of the American Statistical Association},
  volume  = {102},
  number  = {477},
  pages   = {359--378},
  year    = {2007},
  doi     = {10.1198/016214506000001437}
}

@inproceedings{guo2017calibration,
  title     = {On Calibration of Modern Neural Networks},
  author    = {Guo, Chuan and Pleiss, Geoff and Sun, Yu and Weinberger, Kilian Q.},
  booktitle = {Proceedings of the 34th International Conference on Machine Learning},
  series    = {Proceedings of Machine Learning Research},
  volume    = {70},
  pages     = {1321--1330},
  publisher = {PMLR},
  year      = {2017},
  url       = {https://proceedings.mlr.press/v70/guo17a.html}
}

@inproceedings{zheng2023judging,
  title     = {Judging {LLM}-as-a-Judge with {MT-Bench} and {Chatbot Arena}},
  author    = {Zheng, Lianmin and Chiang, Wei-Lin and Sheng, Ying and Zhuang, Siyuan and Wu, Zhanghao and Zhuang, Yonghao and Lin, Zi and Li, Zhuohan and Li, Dacheng and Xing, Eric P. and Zhang, Hao and Gonzalez, Joseph E. and Stoica, Ion},
  booktitle = {Advances in Neural Information Processing Systems (Datasets and Benchmarks Track)},
  volume    = {36},
  pages     = {46595--46623},
  year      = {2023},
  url       = {https://proceedings.neurips.cc/paper_files/paper/2023/hash/91f18a1287b398d378ef22505bf41832-Abstract-Datasets_and_Benchmarks.html}
}

@inproceedings{liu2023geval,
  title     = {{G-Eval}: {NLG} Evaluation using {GPT-4} with Better Human Alignment},
  author    = {Liu, Yang and Iter, Dan and Xu, Yichong and Wang, Shuohang and Xu, Ruochen and Zhu, Chenguang},
  booktitle = {Proceedings of the 2023 Conference on Empirical Methods in Natural Language Processing},
  pages     = {2511--2522},
  year      = {2023},
  address   = {Singapore},
  publisher = {Association for Computational Linguistics},
  doi       = {10.18653/v1/2023.emnlp-main.153},
  url       = {https://aclanthology.org/2023.emnlp-main.153/}
}

@inproceedings{bavaresco2024llms,
  title     = {{LLMs} instead of Human Judges? {A} Large Scale Empirical Study across 20 {NLP} Evaluation Tasks},
  author    = {Bavaresco, Anna and Bernardi, Raffaella and Bertolazzi, Leonardo and Elliott, Desmond and Fern{\'a}ndez, Raquel and Gatt, Albert and Ghaleb, Esam and Giulianelli, Mario and Hanna, Michael and Koller, Alexander and Martins, Andr{\'e} F. T. and Mondorf, Philipp and Neplenbroek, Vera and Pezzelle, Sandro and Plank, Barbara and Schlangen, David and Suglia, Alessandro and Surikuchi, Aditya K. and Takmaz, Ece and Testoni, Alberto},
  booktitle = {Proceedings of the 63rd Annual Meeting of the Association for Computational Linguistics (Volume 2: Short Papers)},
  pages     = {238--255},
  year      = {2025},
  address   = {Vienna, Austria},
  publisher = {Association for Computational Linguistics},
  doi       = {10.18653/v1/2025.acl-short.20},
  url       = {https://aclanthology.org/2025.acl-short.20/}
}

@inproceedings{hada2024metal,
  title     = {{METAL}: Towards Multilingual Meta-Evaluation},
  author    = {Hada, Rishav and Gumma, Varun and Ahmed, Mohamed and Bali, Kalika and Sitaram, Sunayana},
  booktitle = {Findings of the Association for Computational Linguistics: NAACL 2024},
  pages     = {2280--2298},
  year      = {2024},
  address   = {Mexico City, Mexico},
  publisher = {Association for Computational Linguistics},
  doi       = {10.18653/v1/2024.findings-naacl.148},
  url       = {https://aclanthology.org/2024.findings-naacl.148/}
}

@article{qwen3,
  title   = {{Qwen3} Technical Report},
  author  = {{Qwen Team}},
  journal = {arXiv preprint arXiv:2505.09388},
  year    = {2025},
  url     = {https://arxiv.org/abs/2505.09388}
}

@misc{llama32,
  title        = {{Llama 3.2}: Revolutionizing Edge {AI} and Vision with Open, Customizable Models},
  author       = {{Meta}},
  year         = {2024},
  howpublished = {Blog post, September 25, 2024},
  url          = {https://ai.meta.com/blog/llama-3-2-connect-2024-vision-edge-mobile-devices/}
}

@article{gemma3,
  title   = {{Gemma 3} Technical Report},
  author  = {{Gemma Team}},
  journal = {arXiv preprint arXiv:2503.19786},
  year    = {2025},
  url     = {https://arxiv.org/abs/2503.19786}
}

@inproceedings{pombal2025mprometheus,
  title     = {{M-Prometheus}: A Suite of Open Multilingual {LLM} Judges},
  author    = {Pombal, Jos{\'e} and Yoon, Dongkeun and Fernandes, Patrick and Wu, Ian and Kim, Seungone and Rei, Ricardo and Neubig, Graham and Martins, Andr{\'e} F. T.},
  booktitle = {Second Conference on Language Modeling (COLM)},
  year      = {2025},
  url       = {https://openreview.net/forum?id=Atyk8lnIQQ}
}

@inproceedings{kim2024prometheus2,
  title     = {{Prometheus 2}: An Open Source Language Model Specialized in Evaluating Other Language Models},
  author    = {Kim, Seungone and Suk, Juyoung and Longpre, Shayne and Lin, Bill Yuchen and Shin, Jamin and Welleck, Sean and Neubig, Graham and Lee, Moontae and Lee, Kyungjae and Seo, Minjoon},
  booktitle = {Proceedings of the 2024 Conference on Empirical Methods in Natural Language Processing},
  pages     = {4334--4353},
  year      = {2024},
  address   = {Miami, Florida, USA},
  publisher = {Association for Computational Linguistics},
  doi       = {10.18653/v1/2024.emnlp-main.248},
  url       = {https://aclanthology.org/2024.emnlp-main.248/}
}

@inproceedings{joshi2017triviaqa,
  title     = {{TriviaQA}: A Large Scale Distantly Supervised Challenge Dataset for Reading Comprehension},
  author    = {Joshi, Mandar and Choi, Eunsol and Weld, Daniel and Zettlemoyer, Luke},
  booktitle = {Proceedings of the 55th Annual Meeting of the Association for Computational Linguistics (Volume 1: Long Papers)},
  pages     = {1601--1611},
  year      = {2017},
  address   = {Vancouver, Canada},
  publisher = {Association for Computational Linguistics},
  doi       = {10.18653/v1/P17-1147},
  url       = {https://aclanthology.org/P17-1147/}
}

@inproceedings{hendrycks2021mmlu,
  title     = {Measuring Massive Multitask Language Understanding},
  author    = {Hendrycks, Dan and Burns, Collin and Basart, Steven and Zou, Andy and Mazeika, Mantas and Song, Dawn and Steinhardt, Jacob},
  booktitle = {International Conference on Learning Representations (ICLR)},
  year      = {2021},
  url       = {https://openreview.net/forum?id=d7KBjmI3GmQ}
}

@article{dang2024ayaexpanse,
  title   = {{Aya Expanse}: Combining Research Breakthroughs for a New Multilingual Frontier},
  author  = {Dang, John and Singh, Shivalika and D'souza, Daniel and Ahmadian, Arash and Salamanca, Alejandro and Smith, Madeline and Peppin, Aidan and Hong, Sungjin and Govindassamy, Manoj and Zhao, Terrence and Kublik, Sandra and Amer, Meor and Aryabumi, Viraat and Campos, Jon Ander and Tan, Yi-Chern and Kocmi, Tom and Strub, Florian and Grinsztajn, Nathan and Flet-Berliac, Yannis and Locatelli, Acyr and Lin, Hangyu and Talupuru, Dwarak and Venkitesh, Bharat and Cairuz, David and Yang, Bowen and Chung, Tim and Ko, Wei-Yin and Shi, Sylvie Shang and Shukayev, Amir and Bae, Sammie and Piktus, Aleksandra and Castagn{\'e}, Roman and Cruz-Salinas, Felipe and Kim, Eddie and Crawhall-Stein, Lucas and Morisot, Adrien and Roy, Sudip and Blunsom, Phil and Zhang, Ivan and Gomez, Aidan and Frosst, Nick and Fadaee, Marzieh and Ermis, Beyza and {\"U}st{\"u}n, Ahmet and Hooker, Sara},
  journal = {arXiv preprint arXiv:2412.04261},
  year    = {2024},
  url     = {https://arxiv.org/abs/2412.04261}
}

@inproceedings{seabold2010statsmodels,
  title     = {{Statsmodels}: Econometric and Statistical Modeling with {Python}},
  author    = {Seabold, Skipper and Perktold, Josef},
  booktitle = {Proceedings of the 9th Python in Science Conference},
  pages     = {92--96},
  year      = {2010},
  doi       = {10.25080/Majora-92bf1922-011},
  url       = {https://proceedings.scipy.org/articles/Majora-92bf1922-011}
}

@inproceedings{joulin2017bag,
  title     = {Bag of Tricks for Efficient Text Classification},
  author    = {Joulin, Armand and Grave, Edouard and Bojanowski, Piotr and Mikolov, Tomas},
  booktitle = {Proceedings of the 15th Conference of the {E}uropean Chapter of the Association for Computational Linguistics: Volume 2, Short Papers},
  pages     = {427--431},
  year      = {2017},
  address   = {Valencia, Spain},
  publisher = {Association for Computational Linguistics},
  url       = {https://aclanthology.org/E17-2068/}
}

@article{joulin2016fasttextzip,
  title   = {{FastText.zip}: Compressing Text Classification Models},
  author  = {Joulin, Armand and Grave, Edouard and Bojanowski, Piotr and Douze, Matthijs and J{\'e}gou, Herv{\'e} and Mikolov, Tomas},
  journal = {arXiv preprint arXiv:1612.03651},
  year    = {2016},
  url     = {https://arxiv.org/abs/1612.03651}
}

@inproceedings{marchisio2024understanding,
  title     = {Understanding and Mitigating Language Confusion in {LLMs}},
  author    = {Marchisio, Kelly and Ko, Wei-Yin and B{\'e}rard, Alexandre and Dehaze, Th{\'e}o and Ruder, Sebastian},
  booktitle = {Proceedings of the 2024 Conference on Empirical Methods in Natural Language Processing},
  pages     = {6653--6677},
  year      = {2024},
  address   = {Miami, Florida, USA},
  publisher = {Association for Computational Linguistics},
  doi       = {10.18653/v1/2024.emnlp-main.380},
  url       = {https://aclanthology.org/2024.emnlp-main.380/}
}

@inproceedings{etxaniz2024multilingual,
  title     = {Do Multilingual Language Models Think Better in {E}nglish?},
  author    = {Etxaniz, Julen and Azkune, Gorka and Soroa, Aitor and Lopez de Lacalle, Oier and Artetxe, Mikel},
  booktitle = {Proceedings of the 2024 Conference of the North American Chapter of the Association for Computational Linguistics: Human Language Technologies (Volume 2: Short Papers)},
  pages     = {550--564},
  year      = {2024},
  address   = {Mexico City, Mexico},
  publisher = {Association for Computational Linguistics},
  doi       = {10.18653/v1/2024.naacl-short.46},
  url       = {https://aclanthology.org/2024.naacl-short.46/}
}

@inproceedings{artetxe2020translation,
  title     = {Translation Artifacts in Cross-lingual Transfer Learning},
  author    = {Artetxe, Mikel and Labaka, Gorka and Agirre, Eneko},
  booktitle = {Proceedings of the 2020 Conference on Empirical Methods in Natural Language Processing (EMNLP)},
  pages     = {7674--7684},
  year      = {2020},
  address   = {Online},
  publisher = {Association for Computational Linguistics},
  doi       = {10.18653/v1/2020.emnlp-main.618},
  url       = {https://aclanthology.org/2020.emnlp-main.618/}
}

@inproceedings{singh2025global,
  title     = {{Global MMLU}: Understanding and Addressing Cultural and Linguistic Biases in Multilingual Evaluation},
  author    = {Singh, Shivalika and Romanou, Angelika and Fourrier, Cl{\'e}mentine and Adelani, David Ifeoluwa and Ngui, Jian Gang and Vila-Suero, Daniel and Limkonchotiwat, Peerat and Marchisio, Kelly and Leong, Wei Qi and Susanto, Yosephine and Ng, Raymond and Longpre, Shayne and Ruder, Sebastian and Ko, Wei-Yin and Bosselut, Antoine and Oh, Alice and Martins, Andr{\'e} F. T. and Choshen, Leshem and Ippolito, Daphne and Ferrante, Enzo and Fadaee, Marzieh and Ermis, Beyza and Hooker, Sara},
  booktitle = {Proceedings of the 63rd Annual Meeting of the Association for Computational Linguistics (Volume 1: Long Papers)},
  pages     = {18761--18799},
  year      = {2025},
  address   = {Vienna, Austria},
  publisher = {Association for Computational Linguistics},
  doi       = {10.18653/v1/2025.acl-long.919},
  url       = {https://aclanthology.org/2025.acl-long.919/}
}

@inproceedings{ahuja2022calibration,
  title     = {On the Calibration of Massively Multilingual Language Models},
  author    = {Ahuja, Kabir and Sitaram, Sunayana and Dandapat, Sandipan and Choudhury, Monojit},
  booktitle = {Proceedings of the 2022 Conference on Empirical Methods in Natural Language Processing},
  pages     = {4310--4323},
  year      = {2022},
  address   = {Abu Dhabi, United Arab Emirates},
  publisher = {Association for Computational Linguistics},
  doi       = {10.18653/v1/2022.emnlp-main.290},
  url       = {https://aclanthology.org/2022.emnlp-main.290/}
}

@inproceedings{krause2023confidently,
  title     = {Confidently Wrong: Exploring the Calibration and Expression of (Un)Certainty of Large Language Models in a Multilingual Setting},
  author    = {Krause, Lea and Tufa, Wondimagegnhue Tsegaye and B{\'a}ez Santamar{\'\i}a, Selene and Daza, Angel and Khurana, Urja and Vossen, Piek},
  booktitle = {Proceedings of the Workshop on Multimodal, Multilingual Natural Language Generation and Multilingual WebNLG Challenge (MM-NLG 2023)},
  pages     = {1--9},
  year      = {2023},
  address   = {Prague, Czech Republic},
  publisher = {Association for Computational Linguistics},
  url       = {https://aclanthology.org/2023.mmnlg-1.1/}
}

@inproceedings{fu2025reliable,
  title     = {How Reliable is Multilingual {LLM}-as-a-Judge?},
  author    = {Fu, Xiyan and Liu, Wei},
  booktitle = {Findings of the Association for Computational Linguistics: EMNLP 2025},
  pages     = {11040--11053},
  year      = {2025},
  address   = {Suzhou, China},
  publisher = {Association for Computational Linguistics},
  doi       = {10.18653/v1/2025.findings-emnlp.587},
  url       = {https://aclanthology.org/2025.findings-emnlp.587/}
}

@article{fu2026languages,
  title   = {When Languages Disagree: Self-Evolving Multilingual {LLM} Judges},
  author  = {Fu, Xiyan and Lu, Wei},
  journal = {arXiv preprint arXiv:2606.08092},
  year    = {2026},
  url     = {https://arxiv.org/abs/2606.08092}
}

@article{yin2026judge,
  title   = {Does the Judge Prefer {E}nglish? {E}valuating Language-Switching Invariance in {LLM}-as-a-Judge},
  author  = {Yin, Shaojie},
  journal = {arXiv preprint arXiv:2606.14278},
  year    = {2026},
  url     = {https://arxiv.org/abs/2606.14278}
}

@article{mahmood2026rank,
  title   = {Rank Reversal in Multilingual {LLM} Judges: A Label-Free Double-Centering Calibrator},
  author  = {Mahmood, Alhasan and Abdaljalil, Samir and Kurban, Hasan},
  journal = {arXiv preprint arXiv:2608.22432},
  year    = {2026},
  url     = {https://arxiv.org/abs/2608.22432}
}

@inproceedings{behzad2024ask,
  title     = {To Ask {LLM}s about {E}nglish Grammaticality, Prompt Them in a Different Language},
  author    = {Behzad, Shabnam and Zeldes, Amir and Schneider, Nathan},
  booktitle = {Findings of the Association for Computational Linguistics: EMNLP 2024},
  pages     = {15622--15634},
  year      = {2024},
  address   = {Miami, Florida, USA},
  publisher = {Association for Computational Linguistics},
  doi       = {10.18653/v1/2024.findings-emnlp.916},
  url       = {https://aclanthology.org/2024.findings-emnlp.916/}
}

@misc{anthropic2026opus47,
  title        = {System Card: {Claude Opus 4.7}},
  author       = {{Anthropic}},
  year         = {2026},
  howpublished = {April 16, 2026},
  note         = {\url{https://www.anthropic.com/claude-opus-4-7-system-card}},
  url          = {https://www.anthropic.com/claude-opus-4-7-system-card}
}

\appendix
\section{Prompts and Data Provenance}
\label{app:prompts}
These are the exact English instruction strings. The other languages
use fixed localized templates rather than translation at evaluation
time; all six-language templates are in the supplementary data, and
the judge prompt follows below.

\paragraph{LSQA generation directive.}
``Please answer the following question concisely in English.''
The question follows after a blank line in the same user message.

\paragraph{TriviaQA generation directive.}
``Please answer in English. Give a concise answer (1-3 words if
possible).'' The question follows after a blank line.

\paragraph{MMMLU system directive.}
``Select the correct option. Start with exactly one letter A, B,
C, or D, then give a brief justification in English.'' The final
language name is replaced with the requested response language;
the user message contains the localized question and choices.

\paragraph{English verifier template.}
\begin{quote}\small
Question: \{question\}\\
Proposed answer: \{answer\}\\
Is the proposed answer correct? Answer only ``True'' or ``False''.\\
Answer:
\end{quote}
Only the wrapper language changes across $V$; the question and
response placeholders retain their original text. Exact token
IDs and chat-template settings are stored in each verifier
manifest, not assumed identical across tokenizers.

\paragraph{Label tokens.} We tokenized the labels with the ten
tokenizers at the revisions used. ``True'' and ``False'' are single
tokens in all ten, with and without a leading space. Among the most
direct translations, only the Chinese pair is a single token per word
in most tokenizers, and only without a leading space (nine of ten).
For French, Japanese, Korean and Hindi, at least one word of the pair
spans several tokens in every tokenizer without a leading space and in
at least eight with one. A first-token probability would then score a
prefix: one letter for French in nine tokenizers, the first character
for Hindi in eight, and the first syllable of the Korean word for
``false'' in nine. The Korean word for ``no'' and the Hindi words for
``yes'' and ``no'' are also split in nearly all tokenizers. The
supplementary data give the counts for every tokenizer and label
pair.

\paragraph{Judge prompt.} The judge receives one user message
(braces mark the filled fields):
\begin{quote}\footnotesize
\#\#\#Task Description: An instruction (might include an Input inside
it), a response to evaluate, a reference answer that gets a score of 5,
and a score rubric representing an evaluation criterion are given.\\
1. Write detailed feedback assessing the response strictly against the
score rubric, not in general.\\
2. Then write one integer score between 1 and 5, referring to the
score rubric.\\
3. The output format must be exactly: ``Feedback: (feedback for the
criterion) [RESULT] (an integer between 1 and 5)''.\\
4. Do not generate any other opening, closing, markdown, or
explanation.\\[3pt]
\#\#\#The instruction to evaluate:\\
Judge only the factual correctness of the candidate answer to this
multilingual question.\\
Question language: \{Q\}\\
Requested response language: \{R\}\\
Localized question:\\
\{question in $Q$, with choices for MMMLU\}\\[3pt]
\#\#\#Response to evaluate:\\
\{response\}\\[3pt]
\#\#\#Reference Answer (Score 5):\\
Response-language gold (\{R\}): \{reference answer in $R$\}\\
English cross-check (en): \{English reference answer\}\\
English aliases: \{aliases, or ``None supplied''\}\\[3pt]
\#\#\#Score Rubrics: [Multilingual factual correctness]\\
Judge the central fact asked by the localized question. Surface-form
differences, aliases, transliterations, code-switching, a different
answer language, grammatical particles, honorifics, accents,
capitalization, punctuation, and omitted supplementary details do not
reduce the score when the same core answer is conveyed. A bare correct
entity is sufficient. Dates, numbers, or details must match when they
are the fact explicitly asked for. Wrong entities or central facts,
contradictions, hallucinations, refusals, empty answers, generic
deflections, and garbled non-answers are incorrect. Use the middle
score only for genuine ambiguity, unclear partial credit, unreadable
language, or a cut-off answer; default to a decisive score.\\
Score 1: Definitely incorrect: wholly wrong, contradictory,
hallucinated, or a non-answer.\\
Score 2: Incorrect: the candidate does not convey the required central
fact.\\
Score 3: Genuinely uncertain or ambiguous between correct and
incorrect.\\
Score 4: Correct core fact with only negligible, irrelevant, or
supplementary flaws.\\
Score 5: Clearly correct and unambiguously equivalent to the reference
answer.\\[3pt]
\#\#\#Feedback:
\end{quote}
The parser accepts nonempty feedback with an optional
\texttt{Feedback:} prefix followed by a terminal \texttt{[RESULT]} and
a single integer 1--5; any other output is malformed.

\paragraph{Data provenance.} The release manifest records the
selected MMMLU source keys and the exclusion rules with their counts. For TriviaQA
we do not apply a four-token answer filter, and the release includes
longer canonical answers. Every slice is tied to an immutable model
revision and its recorded runtime settings.

\section{Human Review Details}
\label{app:human}
The primary review sample was frozen from the completed set of generated responses
without using judge scores or completion status, so judge-conditioned
convenience sampling plays no part in it. Reviewer packets omit model
identities, judge outputs and confidence scores, but a response can
still reveal its model in its own text, so blinding is not guaranteed.
Reviewers could mark a case unjudgeable rather than infer unfamiliar
language content. The item bootstrap respects the
dependence between repeated configurations of an item but is not an
exact design-based interval.

Same-language units of the translation/reference audit assess
answerability and reference quality rather than cross-language
preservation. Of the 972 sampled responses analyzed,
61.4\% were
judged compliant with the requested response language, 15.4\% were
in another language, 9.5\% mixed languages, 12.7\% were
language-indeterminate (e.g., an option letter alone) and 1.0\% were
empty or refusals (Table~\ref{tab:compliance}).
For a panel-wide check we labeled the realized language of all
1,036,800 responses automatically, with a rule fixed before any
comparison with the reviewers. Each line or sentence takes its
language from its script (Hangul, kana, Han, Devanagari) or, for Latin
script, from fastText language identification
\citep{joulin2017bag,joulin2016fasttextzip}; short Latin strings and
short Han-only strings are neutral, and a response is compliant when
at least 75\% of its non-neutral letters are in the requested
language. On the 972 reviewed responses, the compliant-versus-other
labels agree with the reviewers on 85.6\% (Cohen's $\kappa=0.69$;
precision 0.86, recall 0.92). The rule rarely detects mixing, which
the reviewers found in 9.5\% of responses.

\section{Robustness Details}
\label{app:robustness}
All values in this section come from a supplementary analysis that
reads the same inputs with the same metric and mixed-model code
as the main pipeline. On the full data it reproduces the
overview metrics of all three datasets (Table~\ref{tab:overview}) and the
LSQA and TriviaQA matching coefficients and Wald intervals (Table~\ref{tab:e9}); its
model-resampling stream differs, so we report no resampled sign
fractions here.

\paragraph{Item-level uncertainty.} The intervals come from a
bootstrap over base items with the ten models held fixed.
We resample the 960 items with replacement, within LSQA content
subsets and MMMLU subject groups (TriviaQA is unstratified),
recompute each accuracy contrast from the model-averaged item scores,
and take the 2.5th and 97.5th percentiles of 2,000 replicates. The
intervals therefore reflect item sampling, not variation across
models. Panel accuracy is 0.511 [0.498, 0.524] on LSQA, 0.523
[0.508, 0.539] on MMMLU and 0.316 [0.301, 0.332] on TriviaQA. Of the
contrasts in Sections~\ref{sec:results-mmmlu} and~\ref{sec:results-qr}, only the Korean and Hindi LSQA
native advantages have intervals that include zero.

\paragraph{Judge fallbacks.} Before binary resolution, 8,026 score-3
and 13,872 malformed LSQA judgments (6.3\%) and 4,128 and 2,549
TriviaQA judgments (1.9\%) were neither 4--5 nor 1--2. Their share
ranges over models from 2.5\% to 11.7\% on LSQA and from 1.1\% to
4.5\% on TriviaQA, with Qwen2.5-14B and Mistral-7B highest, and is
largest for requests for Hindi responses (11.4\% and 3.2\%, against
at most 7.3\% and 2.2\% for the other requested languages).
Excluding the responses whose judge
output was score~3 or malformed keeps 323,702 LSQA (93.7\%) and
338,923 TriviaQA (98.1\%) responses. Accuracy rises from 0.511 to
0.540 on LSQA and from 0.316 to 0.322 on TriviaQA, ECE falls by 0.009
and 0.004, and AUROC changes by less than 0.003. The accuracy and
AUROC orderings of the LSQA regimes (Table~\ref{tab:lsqa-regimes})
and of the four TriviaQA English-position groups are unchanged, as
are the signs and interval conclusions of all AUROC matching effects
and of the TriviaQA ECE effects. The LSQA $Q{=}R$ ECE coefficient
falls to 0.009 and its interval then includes zero. Counting score~3
as correct instead gives accuracies of 0.534 (LSQA) and 0.328
(TriviaQA). The human review suggests that neither treatment is
exact: reviewers judged most malformed cases incorrect but most
score-3 cases correct (Section~\ref{sec:human}).

\paragraph{MMMLU responses without a final option.} Excluding the
5,237 MMMLU responses (1.5\%) without a final option raises
accuracy to 0.530 and lowers ECE to 0.367, with AUROC 0.688. The
question-language pattern is unchanged (0.626 for English questions,
0.431 for Hindi) and the requested-response-language range stays
narrow (0.522--0.547). The AUROC matching effects also remain:
$Q{=}R$ is 0.006 [0.002, 0.010] and $R{=}V$ 0.005, both intervals
excluding zero, so the small MMMLU AUROC effects in Table~\ref{tab:e9}
do not depend on responses without a final answer.

\paragraph{Without Qwen3-8B.} Qwen3-8B is the only model whose
responses were regenerated after the human review. Dropping it changes
panel accuracy to 0.512 (LSQA), 0.519 (MMMLU) and 0.322 (TriviaQA).
All LSQA and TriviaQA matching coefficients keep their signs and
interval conclusions, and the LSQA regime and TriviaQA group orderings
are unchanged. On MMMLU the English--Hindi question-language
difference is 0.197 [0.182, 0.213], English responses remain 0.022
[0.019, 0.025] above the non-English mean, and $Q{=}R$ and $R{=}V$
still raise AUROC slightly (0.006 and 0.005, intervals excluding
zero). The Hindi native advantage on LSQA becomes $-0.031$ [$-0.064$,
$-0.0004$], an interval that excludes zero.

\paragraph{Truncated inputs.} Rebuilding every prompt with the
recorded tokenizer and chat template, which reproduces all recorded
prompt hashes, shows that 174 generation and 1,939 verifier prompts
(0.02\%--0.03\%) exceeded the 2,048-token limit and lost their final
turn. The 357 affected MMMLU responses come from 16 items, asked
mostly in Hindi, in college medicine, high-school European and US
history and professional law; the Gemma tokenizers keep every prompt within the
limit. Excluding them moves MMMLU accuracy from 0.5232 to 0.5234,
changes AUROC (0.6908) and ECE (0.3687) by less than 0.0001, and
leaves every matching-effect interval conclusion unchanged.

\paragraph{Realized response language.} The automatic labels of
Appendix~\ref{app:human} mark 74.1\% of LSQA, 63.2\% of MMMLU and
51.9\% of TriviaQA responses as written in the requested language;
TriviaQA has the most language-neutral short answers. Keeping only
these responses gives accuracies of 0.516, 0.502 and 0.297. All LSQA
and TriviaQA AUROC matching effects keep their interval conclusions
(only the null $Q{=}V$ terms change sign); the LSQA regime accuracies keep their order, and AUROC
stays highest on the off-native diagonal and lowest when only the
question matches. The calibration costs of matching grow: the
$Q{=}R$ ECE coefficient becomes 0.025 on LSQA and 0.033 on TriviaQA.
On MMMLU the AUROC matching effects become null ($Q{=}R$ 0.008
[$-0.003$, 0.018]), and question language still separates accuracy
far more than requested response language (a range of 0.217 across
question languages against 0.089 across requested languages). On
TriviaQA the English-question advantage falls from 0.213 to 0.152
for English responses and from 0.117 to 0.081 for other responses,
while the English-response advantage rises from 0.179 to 0.206 for
English questions and from 0.083 to 0.135 for other questions. Part
of the full-data English-question advantage thus comes from answers
given in English despite a request for another language. Compliance
is itself an outcome of the requested language and the model, so
these are conditional descriptions rather than effects.

The remaining paragraphs of this appendix come from a second
supplementary analysis that reads the main pipeline's output tables and the locked
human review.

\paragraph{Model-level test of matching effects.} Fitting the matching
regression separately for each model and testing the mean of the ten
coefficients ($t$, 9 df) confirms the AUROC effects of $Q{=}R$ and
$R{=}V$ on LSQA ($t=4.71$ and 4.85; nine models share the sign) and
TriviaQA ($t=3.72$ and 4.20; nine and ten models) and the negative
three-way term ($t=-4.28$ and $-3.71$). On MMMLU the same AUROC
effects are not distinguishable from zero ($t=1.54$ and 2.24; seven
and eight models). No ECE effect of $Q{=}R$ or $R{=}V$ on LSQA or
TriviaQA reaches significance ($|t|\le 2.07$; six to eight models),
whereas $Q{=}V$ and the three-way match lower TriviaQA ECE ($t=-2.70$
and $-2.91$) and $R{=}V$ lowers MMMLU ECE ($t=-2.64$; nine models).
The default optimizer of the mixed model in Table~\ref{tab:e9} stops at a singular
fit for LSQA and TriviaQA AUROC; its matching coefficients equal the
model-averaged per-model coefficients above.

\paragraph{Judge error by language and model.} On the 316 judgeable
LSQA review responses, the judge's correct rate minus the reviewers'
is 0.000 for requested English and Korean, $-0.019$ for Japanese,
$-0.075$ for French, $-0.113$ for Chinese and $-0.140$ for Hindi
(about 50 responses each; bootstrap intervals span about
$\pm$0.1); on TriviaQA it ranges from $-0.019$ to $-0.113$. On both
benchmarks agreement is 0.915 [0.858, 0.962] when question and
response share a language and 0.845 [0.815, 0.876] otherwise. By
model, the judge under-credits its own base model, Qwen2.5-14B, the
most ($-0.200$ [$-0.300$, $-0.100$]; other models $-0.127$ to
$+0.042$).

\paragraph{Content alignment with language strength held fixed.} For
each model we regress LSQA accuracy on the 216 subset--$Q$--$R$ cells
with subset effects, the full $Q{\times}R$ table and subset-specific
indicators for $Q{=}s$ only, $R{=}s$ only and $Q{=}R{=}s$. The
$Q{=}R{=}s$ gain (mean over models; $t$ interval, 9 df) is $-0.050$
[$-0.088$, $-0.012$] on English, 0.052 [0.012, 0.091] on French,
0.102 [0.072, 0.132] on Chinese, 0.094 [0.052, 0.137] on Japanese,
0.175 [0.159, 0.190] on Korean and 0.189 [0.127, 0.251] on Hindi; the
last four are positive for all ten models, and question-only and
response-only alignment also help on these subsets (0.06--0.14). The
unadjusted native advantage is small on Korean and Hindi because their
unaligned conditions include English questions and responses, which
raise accuracy in general.

\paragraph{Pooled versus cell-level regime metrics.} Table~\ref{tab:lsqa-regimes}
pools each model--subset--$V$ unit's responses across the regime's
$Q/R$ conditions, so AUROC can also reward differences between
conditions in regimes that pool several of them. Averaging the
160-item configuration cells instead gives AUROC 0.797, 0.795, 0.783,
0.805 and 0.788 and ECE 0.264, 0.264, 0.268, 0.293 and 0.281 for the
five regimes in table order; accuracy is unchanged.

\section{Thinking-Template Ablation}
\label{app:thinking}
For Qwen3-8B on LSQA with English verifier instructions (pooled over
its 34,560 responses), reading the verifier logits under the
thinking-enabled template leaves the responses and accuracy unchanged
(0.502) but moves the median $\ptrue$ from 1.000 to 0.516. ECE
falls (0.236 to 0.143), but the Brier score rises (0.238 to 0.270)
and discrimination drops close to chance (AUROC 0.862 to 0.572).
This is a template-mode sensitivity of the direct probe
(Section~\ref{sec:confidence}), not evidence about reasoning.

\section{Additional Results}
\label{app:additional}
\begin{figure*}[t]
\centering
\includegraphics[width=0.8\textwidth]{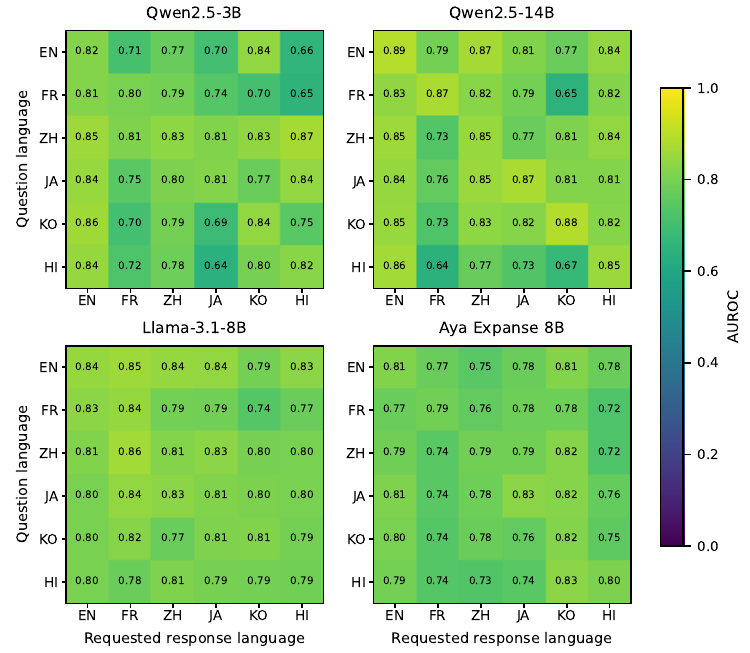}
\caption{LSQA AUROC by question language (rows) and requested
response language (columns) for four models fixed before the analysis
(the smallest and largest Qwen2.5 models, Llama-3.1-8B and Aya
Expanse 8B), with English verifier instructions, pooled across content
subsets. The supplementary data give the cells of all ten models.}
\label{fig:heatmaps}
\end{figure*}

Figure~\ref{fig:heatmaps} redraws the main pipeline's figure source
tables with the notation of the main text, as Figure~\ref{fig:summary}
does in Section~\ref{sec:results-qr}.

\begin{table}[t]
\centering
\small
\setlength{\tabcolsep}{4pt}
\begin{tabular}{lcccc}
\toprule
Subject group & $Q,R$=EN & $Q$=EN & $R$=EN & Neither \\
\midrule
Humanities & .581 & .562 & .463 & .442 \\
STEM & .581 & .569 & .505 & .482 \\
Social sciences & .708 & .684 & .561 & .538 \\
Other & .668 & .653 & .560 & .539 \\
\bottomrule
\end{tabular}
\caption{MMMLU exact-match accuracy by subject group and English
position. $Q$=EN: English question with a non-English requested
response; $R$=EN: the converse.}
\label{tab:mmmlu-groups}
\end{table}

\begin{table}[t]
\centering
\small
\begin{tabular}{lcc}
\toprule
Model & Exact match & Judge \\
\midrule
Qwen2.5-3B & .484 & .551 \\
Qwen2.5-7B & .565 & .729 \\
Qwen2.5-14B & .653 & .838 \\
Qwen3-8B & .562 & .741 \\
Llama-3.1-8B & .523 & .640 \\
Llama-3.2-3B & .449 & .540 \\
Gemma-2-9B & .578 & .740 \\
Gemma-3-12B & .550 & .753 \\
Mistral-7B-v0.3 & .363 & .365 \\
Aya Expanse 8B & .505 & .688 \\
\midrule
Panel & .523 & .658 \\
\bottomrule
\end{tabular}
\caption{MMMLU accuracy over all 34,560 responses per model. The
judge column is the sensitivity analysis, not the primary label.}
\label{tab:mmmlu-models}
\end{table}

\begin{table}[t]
\centering
\small
\setlength{\tabcolsep}{3pt}
\begin{tabular}{lcccc}
\toprule
& Compl. & Non. & Mixed & Indet. \\
\midrule
Qwen2.5-3B & .620 & .157 & .074 & .130 \\
Qwen2.5-7B & .704 & .093 & .037 & .157 \\
Qwen2.5-14B & .639 & .139 & .065 & .139 \\
Llama-3.1-8B & .491 & .306 & .074 & .120 \\
Llama-3.2-3B & .435 & .370 & .056 & .111 \\
Gemma-2-9B & .648 & .037 & .120 & .194 \\
Gemma-3-12B & .620 & .009 & .176 & .194 \\
Mistral-7B-v0.3 & .667 & .139 & .176 & .009 \\
Aya Expanse 8B & .704 & .139 & .074 & .083 \\
\midrule
Requested EN & .340 & .383 & .099 & .179 \\
Requested FR & .599 & .093 & .074 & .210 \\
Requested ZH & .765 & .136 & .031 & .049 \\
Requested JA & .741 & .111 & .086 & .062 \\
Requested KO & .667 & .123 & .142 & .056 \\
Requested HI & .574 & .080 & .136 & .204 \\
\bottomrule
\end{tabular}
\caption{Human-rated realized response language on the 972-response
sample without Qwen3-8B (108 per model, 162 per requested language).
Rows omit the empty/refusal share (1.0\% overall).}
\label{tab:compliance}
\end{table}

\begin{figure*}[t]
\centering
\includegraphics[width=0.94\textwidth]{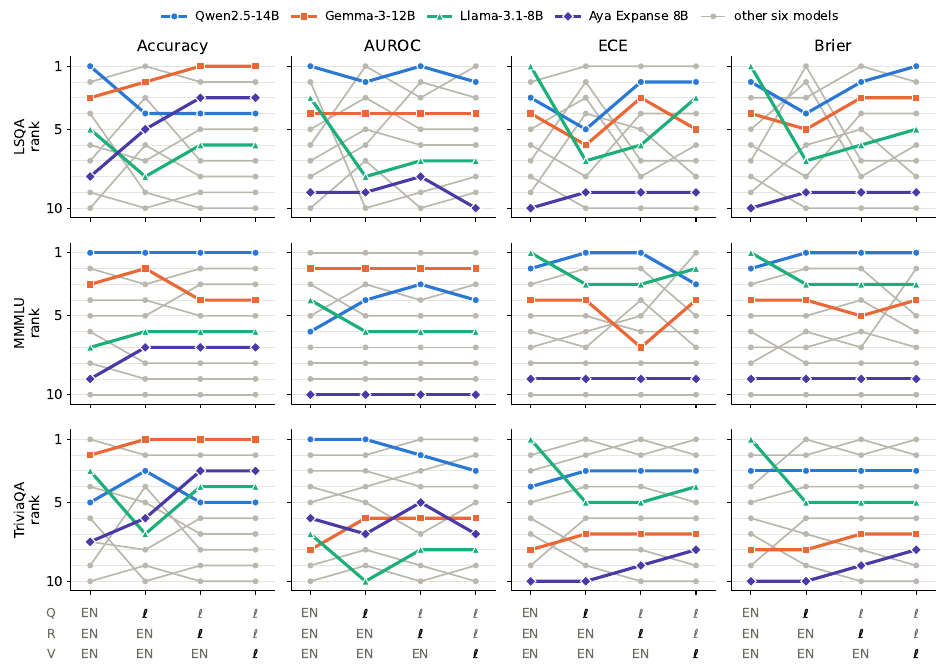}
\caption{Model ranks by benchmark along the path of
Figure~\ref{fig:teaser}, averaged over the five target languages.}
\label{fig:ranks}
\end{figure*}

\paragraph{Model rankings.} Figures~\ref{fig:teaser}
and~\ref{fig:ranks} follow a path from the all-English setting to
the fully localized one that switches one language at a time from
English to a target language $\ell$, the same $\ell$ in every switched
position: $(Q,R,V)$ = \mbox{(EN, EN, EN)}, \mbox{($\ell$, EN, EN)},
\mbox{($\ell$, $\ell$, EN)} and \mbox{($\ell$, $\ell$, $\ell$)}. A model's score in a setting is its
configuration-cell metric averaged over $\ell$ = FR, ZH, JA, KO and HI;
Figure~\ref{fig:teaser} also averages the three benchmarks, and MMMLU
is scored by exact match. Rank intervals and pair comparisons use the item bootstrap of
Figure~\ref{fig:effects} (the same 2,000 draws). A model pair counts
as reversed between two settings when the 95\% interval of the pair's
score difference excludes zero in both settings, with opposite signs;
we do not adjust for the number of pairs. Table~\ref{tab:ranks} gives
these counts and Kendall's $\tau$ between the all-English and fully
localized rankings. Of the 39 accuracy pairs whose order is
significant in the all-English setting, six are reversed in the fully
localized one: Aya Expanse 8B overtakes four models, Gemma-3-12B overtakes
Qwen2.5-14B, and Llama-3.2-3B overtakes Mistral-7B-v0.3. Accuracy
cannot change at the verifier step, but one ECE pair and three Brier
pairs still reverse there. In the off-path setting (EN, EN, $\ell$),
where only the verifier language is switched, six ECE pairs reverse,
and Llama-3.2-3B falls in ECE from second (rank interval 2--3) to
ninth (8--9). The four colored models in Figures~\ref{fig:teaser}
and~\ref{fig:ranks} were picked after seeing the results: the
all-English accuracy leader (Qwen2.5-14B), the fully localized
accuracy leader (Gemma-3-12B), the all-English ECE and Brier leader
(Llama-3.1-8B), and the model whose accuracy rank rises most between
these two settings (Aya Expanse 8B, eighth to fourth); four is the
most line colors that passed our color-vision check. Ranks are over
all ten models, and the supplementary data give every score, rank and
rank interval.

Single configurations also depart from the all-English ranking far
more than sampling noise would. Against the all-English
configuration, the median $\tau$ of the
single configurations ranges over the four metrics from 0.11 to 0.38
on LSQA, from 0.47 to 0.56 on TriviaQA and from 0.60 to 0.73 on
MMMLU, whereas bootstrap resamples of the all-English configuration
itself give a mean $\tau$ of 0.70 to 0.95.

\begin{table}[t]
\centering
\small
\setlength{\tabcolsep}{3.5pt}
\begin{tabular}{lcccccc}
\toprule
& & \multicolumn{5}{c}{Reversed pairs} \\
\cmidrule(l){3-7}
Metric & $\tau$ & $Q$ & $R$ & $V$ & Ends & $V$ only \\
\midrule
Accuracy & .51 [.42, .60] & 8 & 5 & 0 & 6 & 0 \\
AUROC & .73 [.56, .87] & 1 & 0 & 0 & 0 & 0 \\
ECE$_{15}$ & .87 [.64, .91] & 5 & 0 & 1 & 2 & 6 \\
Brier & .73 [.51, .78] & 4 & 0 & 3 & 3 & 0 \\
\bottomrule
\end{tabular}
\caption{Rank changes along the path of Figure~\ref{fig:teaser}
(means over the three benchmarks). $\tau$: Kendall's $\tau$ between
the all-English and fully localized rankings, with 95\% bootstrap
interval. The other columns count reversed model pairs (of 45) at the
$Q$, $R$ and $V$ steps, between the two ends of the path, and when
only $V$ is switched.}
\label{tab:ranks}
\end{table}

\begin{table*}[t]
\centering
\small
\setlength{\tabcolsep}{4pt}
\begin{tabular}{p{3.3cm}p{6.4cm}p{5.4cm}}
\toprule
Phenomenon & Question and response (excerpts) & Labels and $\ptrue$ by verifier language \\
\midrule
Judge over-credits an MMMLU answer (Aya Expanse 8B; $Q$=FR, $R$=EN)
& Q: \textit{Lequel des \'enonc\'es suivants sur le clivage embryonnaire
n'est PAS vrai~?} (gold: C)\newline
R: \textit{La r\'eponse incorrecte est A) La pr\'esence de jaune inhibe le
clivage.} [\ldots]
& Judge score 5 (correct). Reviewer: option A, incorrect; answered in
French. Exact match: incorrect. $\ptrue$: FR 0.002, ZH 0.68, other
$V$ $\geq$0.99. \\
\addlinespace
Correct but not in the requested language (Gemma-2-9B, TriviaQA;
$Q$=EN, $R$=KO)
& Q: \textit{Cobweb is a fairy in which Shakespeare play?}\newline
R: \textit{A Midsummer Night's Dream}
& Judge score 4 (correct). Reviewer: correct; answered in English.
$\ptrue\geq$0.999 for all six $V$. \\
\addlinespace
Verifier-language spread (Qwen2.5-7B, LSQA; $Q$=EN, $R$=FR)
& Q: \textit{When did the Great Fire of London occur?}\newline
R: \textit{Le Grand Incendie de Londres s'est produit du 2 au 6
septembre 1666.}
& Judge score 5 (correct). Reviewer: correct; compliant.
$\ptrue$: EN 1.00, FR 1.00, KO 0.18, JA 0.03, ZH 0.0001,
HI $3\times10^{-8}$. \\
\bottomrule
\end{tabular}
\caption{Human-validated examples. For each of the first two
phenomena we took the first qualifying unit, in random review-ID order,
among reviewed responses (Qwen3-8B excluded) whose question and
response are in Latin script; the last row is the unit with the widest
$\ptrue$ range across verifier languages among the 109 such units whose
reviewer and final labels agree. Examples illustrate phenomena and do not estimate their
prevalence; Markdown is omitted.}
\label{tab:examples}
\end{table*}

\section{Reviewer Instructions}
\label{app:instructions}
The complete instructions follow verbatim. They were written for six
language assignments, which the five authors shared (one author took
two); field names refer to the columns of the rating sheets.

{\small\raggedright\setlength{\parindent}{0pt}\setlength{\parskip}{4pt}%
\renewcommand{\_}{\textunderscore\allowbreak}%
These files are \textbf{unrated templates}. Six reviewers each receive 180 responses and 72 item-quality/translation checks in their assigned language. There is one primary reviewer per unit, not two independent reviewers. Work only on your assigned copy. Use the assigned opaque reviewer ID, not your name. Do not view admin mappings, model names, automated judge output or confidence scores. Do not use another LLM as your human judgment. If the actual response is in a language you cannot assess, abstain and flag for optional competent author review; never guess from the requested language alone. Reference answers can be wrong. English/source-language competence is not assumed.

For each completed unit, fill \texttt{reviewer\_id}, relevant judgment fields and notes, then change \texttt{status} to \texttt{RATED}. Mark \texttt{UNRATED} while unfinished. Use \texttt{unjudgeable}/\texttt{cannot\_assess} for a completed assessment that cannot determine the requested property; these are not incorrect labels. Include a brief reason and any external evidence you actually consulted in \texttt{evidence\_reference}; do not invent evidence. Do not alter question, reference, response, requested language or review ID. A spreadsheet may visually clip long text; inspect the full cell or JSONL text before judging.

\paragraph{Response correctness and language}

\begin{itemize}\setlength{\itemsep}{0pt}
\item \texttt{original\_question\_comprehension}: \texttt{understood\_original}, \texttt{localized\_question\_only}, or \texttt{cannot\_assess}. \texttt{question} is exactly what the model saw (including MMMLU options); \texttt{question\_in\_reviewer\_language} is the processed dataset's localized version. If you cannot understand the original question, you may judge correctness relative to the localized question, but MUST record \texttt{localized\_question\_only}. This does not validate equivalence to the original question. If neither can be assessed, use \texttt{cannot\_assess} and correctness \texttt{unjudgeable}. \texttt{english\_reference\_question} is optional support, not proof that English/source comprehension is available.
\item \texttt{correctness}: \texttt{correct}, \texttt{incorrect}, \texttt{unjudgeable}. Judge whether the response answers the question correctly overall, using the reference as guidance, accepting valid paraphrases/aliases and taking substantive false statements into account. A correct name with a contradictory/wrong explanation is not automatically correct. Ignore style or a language mismatch when assigning factual correctness; record language separately. If the response simply refuses or is empty despite an answerable question, label correctness \texttt{incorrect}. If the question/reference itself is defective and correctness cannot be established, use \texttt{unjudgeable}.
\item \texttt{reference\_issue}: \texttt{none}, \texttt{suspected\_reference\_error}, \texttt{translation\_or\_question\_issue}, \texttt{insufficient\_context}, \texttt{cannot\_assess}; explain concerns instead of forcing reference agreement.
\item MMMLU only: \texttt{mmmlu\_selected\_option} = \texttt{A}, \texttt{B}, \texttt{C}, \texttt{D}, \texttt{multiple}, \texttt{none}, or \texttt{unclear}; \texttt{mmmlu\_option\_correctness} = \texttt{correct}, \texttt{incorrect}, \texttt{unjudgeable}; \texttt{mmmlu\_justification\_correctness} = \texttt{correct}, \texttt{incorrect}, \texttt{unjudgeable}, \texttt{no\_justification}; \texttt{mmmlu\_option\_explanation\_conflict} = \texttt{yes}, \texttt{no}, \texttt{unclear}. Keep option accuracy separate from overall correctness and explanation accuracy. Non-MMMLU rows: enter \texttt{not\_applicable} in these four fields. Do not infer an option from unsupported details if the answer is ambiguous.
\item \texttt{observed\_languages}: comma-separated codes for languages actually used, or \texttt{indeterminate}. Common codes: \texttt{en}, \texttt{fr}, \texttt{zh}, \texttt{ja}, \texttt{ko}, \texttt{hi}; use a language name/code for others.
\item \texttt{language\_status}: \texttt{compliant} (substantive text in requested language), \texttt{noncompliant} (substantive text in another language), \texttt{mixed} (substantial code switching), \texttt{indeterminate} (language-neutral content such as A/B/C/D, numbers, formulae, or short shared names), or \texttt{empty\_or\_refusal}. For a refusal, also record any identifiable language in \texttt{observed\_languages}. A short name or answer letter is not automatically English or compliant. Borrowed names alone do not make otherwise natural text mixed.
\item \texttt{has\_english\_fallback} and \texttt{has\_code\_switching}: \texttt{yes}, \texttt{no}, or \texttt{unclear}. Use \texttt{unclear} when short/neutral text prevents determination; explain partial use of English vs other substantive language in notes.
\end{itemize}

The question may be in one language and the requested answer in another. MMMLU options in the question preserve the model's question-language presentation; localized question and extra reference choices are in the requested response language. An unforeseen third language requires competence or abstention. English TriviaQA aliases are optional supporting information, not exhaustive or mandatory surface forms.

\paragraph{Translation, answer validity and cultural context}

The reference/target pair identifies the comparison. LSQA reference language comes from the recovered native-language declaration; it is not an independently authenticated original. Do not assume English is the cultural original. \texttt{source\_comprehension}: \texttt{understood\_source}, \texttt{same\_language\_reference}, or \texttt{cannot\_assess\_source}. If the source language is unknown or you do not understand it, use \texttt{cannot\_assess\_source} and \texttt{meaning\_preserved=cannot\_assess}; do not infer a translation pass just because the target reads well. You may still evaluate target-language answerability, answer validity and cultural context within your competence, explaining the limitation. A bilingual/source-qualified author can optionally investigate flagged cases later; do not prefill their judgment.

For \texttt{meaning\_preserved}, \texttt{reference\_answer\_valid}, \texttt{cultural\_appropriateness}, and \texttt{answerability}, use \texttt{pass}, \texttt{minor\_issue}, \texttt{major\_issue}, or \texttt{cannot\_assess}. Assess the target question and answer jointly, not just fluent phrasing. \texttt{reference\_answer\_valid} means whether the provided target answer is valid for the target question, taking the source/reference pair and subject knowledge into account. Cultural appropriateness concerns preservation of culture-specific entities, time periods, names and assumptions, not subjective preference. Distinguish stylistic awkwardness from a changed answer or unanswerable question.

\texttt{choice\_alignment}: \texttt{pass}, \texttt{minor\_issue}, \texttt{major\_issue}, \texttt{cannot\_assess}, or \texttt{not\_applicable} (non-MMMLU). In MMMLU verify the same semantic choices occupy the same letters and the answer letter/text matches, not merely that answer letters are identical. \texttt{severity}: \texttt{none}, \texttt{minor}, \texttt{major}, or \texttt{cannot\_assess}. Explain errors precisely and cite verifiable evidence when available.

When reference and target languages coincide, mark \texttt{meaning\_preserved=not\_applicable}; still assess answer validity, answerability, cultural context and MMMLU choice integrity. These are source-quality checks, not interlanguage translation evidence.

\paragraph{Optional author adjudication --- only after primary ratings are locked}

Do not fill the optional adjudication template while making primary ratings. The coordinator preserves the original rating first, then identifies flagged/ambiguous cases and appoints an appropriately qualified author/adjudicator if available. Record the locked primary record, escalation reason, reviewed fields, final outcome and rationale; retain the initial and final label separately. This is not a second independent rating and cannot establish inter-rater agreement. Cases that cannot be resolved remain unjudgeable/cannot\_assess. No automated fallback should force uncertain human judgments to incorrect.\par
}

\section{Human-Validated Examples}
\label{app:examples}
Table~\ref{tab:examples} shows three responses from the human review
sample.

\end{document}